\documentclass[letterpaper]{article}
\usepackage{aaai23}

\usepackage{times}  
\usepackage{helvet}  
\usepackage{courier}  
\usepackage[hyphens]{url}  
\usepackage{graphicx} 
\usepackage{natbib}  
\usepackage{caption} 
\usepackage{algorithm}
\usepackage{algorithmic}

\usepackage{newfloat}
\usepackage{listings}
\DeclareCaptionStyle{ruled}{labelfont=normalfont,labelsep=colon,strut=off} 
\floatstyle{ruled}
\newfloat{listing}{tb}{lst}{}
\floatname{listing}{Listing}

\usepackage{amsmath}
\usepackage{amsfonts}
\usepackage{amssymb}
\usepackage{amsthm}
\usepackage{caption}
\usepackage{subcaption}
\usepackage{cleveref}
\usepackage{booktabs}
\usepackage[svgnames,table]{xcolor}
\usepackage[tableposition=above]{caption}
\usepackage{pifont}
\newcommand{\indep}{\perp \!\!\! \perp}

\newif\ifblackandwhite

\usepackage[hmargin=2cm,vmargin=2.5cm]{geometry}
\usepackage{etoolbox}
\usepackage{longtable}%
\usepackage{amsmath,amsthm}
\usepackage{thmtools}
\usepackage{enumitem}

\usepackage{pdflscape}
\usepackage[svgnames]{xcolor}
\usepackage{colortbl}%
  \newcommand{\myrowcolour}{\rowcolor[gray]{0.925}}
\usepackage{booktabs}

\ifblackandwhite
  \newcommand{\cheading}[2]{\textbf{#1\hfill #2}}
  \newcommand{\highest}[1]{\textbf{#1}}
\else
  \newcommand{\cheading}[2]{\textcolor{Maroon}{\textbf{#1\hfill #2}}}
  \newcommand{\highest}[1]{\textcolor{Maroon}{\textbf{#1}}}%
\fi

\newtheorem{proposition}{Proposition}
\newtheorem{assumption}{Assumption}
\newtheorem{theorem}{Theorem}

\usepackage{lipsum}
\usepackage{graphicx}
\begin{document}
%
\title{Gradient Estimation for Binary Latent Variables via Gradient Variance Clipping}
 \author{Russell Z. Kunes$^{1,2,4}$,  Mingzhang Yin$^{4,5}$, Max Land$^{2}$, Doron Haviv$^{2}$, Dana Pe'er$^{2,3}$, Simon Tavaré$^{1,4}$}
 \affiliations{ $^1$Department of Statistics, Columbia University\\
 $^2$Computational and Systems Biology, Memorial Sloan Kettering Cancer Center\\
 $^3$Howard Hughes Medical Institute\\
 $^4$Irving Institute of Cancer Dynamics, Columbia University \\
 $^5$Warrington College of Business, University of Florida \\}

%

\maketitle
\begin{abstract}
\begin{quote}
Gradient estimation is often necessary for fitting generative models with discrete latent variables, in contexts such as reinforcement learning and variational autoencoder (VAE) training. The DisARM estimator  (Yin et al. 2020; Dong, Mnih, and Tucker 2020) achieves state of the art gradient variance for Bernoulli latent variable models in many contexts. However, DisARM and other estimators have potentially exploding variance near the boundary of the parameter space, where solutions tend to lie. To ameliorate this issue, we propose a new gradient estimator \textit{bitflip}-1  that has lower variance at the boundaries of the parameter space. As bitflip-1 has complementary properties to existing estimators, we introduce an aggregated estimator, \textit{unbiased gradient variance clipping} (UGC) that uses either a bitflip-1 or a DisARM gradient update for each coordinate.  We theoretically prove that UGC has uniformly lower variance than DisARM.
Empirically, we observe that UGC achieves the optimal value of the optimization objectives  in toy experiments, discrete VAE training, and in a best subset selection problem. 
\end{quote}
\end{abstract}

\noindent 
\section*{Introduction} Many modern machine learning tasks rely on stochastic gradient estimators, where the estimand is the gradient of an expected value $\mathbb{E}_{\mathbf{z}\sim p(\mathbf{z};\theta)}[f(\mathbf{z})]$ that is intractable to compute \citep{Mohamed2020}.  For example, in reinforcement learning it is often of interest to compute the gradient of an expected reward with respect to the parameters of a distribution over actions, where the reward may be a black box function of discrete states and actions \cite{li2017deep}. In variational inference, the objective function is the evidence lower bound, expressed as an expected value of the log joint probability of latent variable and data under a variational distribution \cite{ranganath2014black,blei2017variational}. In many cases $\mathbf{z}$ is discrete; for example, in the design of biological sequences \cite{brookes2019conditioning} or in models with spike and slab Bayesian priors \cite{moran2021identifiable}.

When the latent variables $\mathbf{z}$ are discrete and high dimensional, there are several challenges in optimizing the mean-valued objective with respect to the distributional parameters $\theta$. First, computing the exact expectation often requires an intractable number function evaluations due to an exponential number of summation terms \cite{aueb2015}. Moreover, the derivative of the function itself with respect to discrete variables is not well defined so the chain rule-based reparametrization trick \citep{kingma2013auto} cannot be used.

A number of methods for estimating the gradient of expected values with respect to discrete random variables have been devised \citep{dong2020disarm,dimitriev2021arms,dong2021coupled,yin2019arsm,aueb2015local,tucker2017rebar,grathwohl2017backpropagation,titsias2022double}.  
A central role shared among the designs of useful gradient estimation is to control the bias and variance of the estimates. 
One line of research reduces the gradient variance in a trade-off of introducing bias. 
Widely used methods include continuous relaxations such as the Gumbel-softmax trick \cite{jang2016categorical,maddison2016concrete,paulus2020gradient}, and the straight through gradient estimator \cite{bengio2013estimating,yin2019understanding}, which have been successfully applied for learning latent representations of images \cite{razavi2019generating} and text \cite{tran2019discrete}. Another line of work considers unbiased estimates that offer guarantees of convergence under conditions on the learning rate sequence \citep{ranganath2014black,robbins1951stochastic}. 
Some methods construct control variate baselines by continuous relaxation of the discrete distributions \citep{tucker2017rebar,grathwohl2017backpropagation}, by first-order Taylor expansions \citep{gu2015muprop,Titsias2022}, or by Stein operators \citep{Shi2022}. Other methods reduce the estimator variance by applying antithetic sampling and coupled sampling  \citep{yin2018arm,dong2020disarm,dimitriev2021arms,yin2019arsm,yin2020probabilistic,kool2019buy}. Our work proceeds in this direction of designing unbiased and low-variance gradient estimators for discrete optimization.


In this work, we notice that in the context of Bernoulli discrete latent variables, a number of existing unbiased methods have unfavorably high variance at the boundary of the parameter space (namely, near $0$ and near $1$) due to reliance on an importance weight that is necessary in order to maintain unbiasedness. To address this downside of existing estimators, we introduce an \textit{unbiased gradient variance clipping} (UGC) estimator that sidesteps this issue by conditionally using one of two types of gradient estimators. For a given coordinate, when values of the probability parameter $\theta$ are near $\frac{1}{2}$, UGC updates the parameter values in the direction of the DisARM gradient estimate. On the other hand, when values of the probability $\theta$ become close to the boundary, UGC transitions to using a novel gradient estimator, \textit{bitflip}-1, that has \textit{complementary} properties to existing estimators that require $O(1)$ function evaluations. 
Namely, rather than considering coordinate-wise independent samples of $\mathbf{z}$, bitflip-1 updates only a single coordinate of the parameter vector at a time, while holding other coordinates fixed to minimize variance. The result is that bitflip-1 has variance linear in the latent dimension $K$ but without explicit dependence on the latent Bernoulli parameters. Our proposed estimator, UGC, has guaranteed uniformly lower variance than DisARM and is robust across practical problems where either DisARM or bitflip-1 alone may fail.
\section*{Background}
Consider the problem of estimating the gradient: 
\begin{align}
  \nabla_\theta \mathbb{E}_{p(\mathbf{z};\theta)}[f(\mathbf{z})]  \label{eq:obj}
\end{align}
where $\mathbf{z} = (z_1,\dots,z_K)$, $z_i \sim \text{Bernoulli}(\theta_i)$, $\theta_i \in [0,1]$, independently, $p(\mathbf{z};\theta) = \prod_{i=1}^K \text{Bernoulli}(\theta_i)$, and $f$ is a potentially complicated and nonlinear function with domain on the lattice. This problem arises in discrete latent variable modeling and reinforcement learning. To compute the exact gradient, we can replace the expectation in \Cref{eq:obj} with the summation over all possible values of $\mathbf{z}$ which has $2^K$ summation terms. Computing the exact gradient thus requires an exponential number of evaluations which is infeasible to compute per iteration of gradient descent in high dimensional problems. Specifically we focus on the context of Bernoulli VAEs where $p_{\lambda}(\mathbf{x}_i|\mathbf{z}_i)$ is parameterized by a neural network, while $\mathbf{z} \in \{0,1\}^K$, and we fit an encoder network $q_\theta(\mathbf{z}|\mathbf{x})$ to maximize the evidence lower bound (ELBO):
\begin{align*}
    \mathcal{L}(\theta, \lambda) = \mathbb{E}_{q}\big\{ \log p_\lambda(\mathbf{x}|\mathbf{z}) +\log p(\mathbf{z}) - \log q_\theta(\mathbf{z}|\mathbf{x})\big\}.
\end{align*}
The exact gradient of the objective function with respect to $\theta$ involves $2^k$ terms in general. As a result, we are forced to use a stochastic estimate of the gradient. Two methods are commonly applied for this task; score function gradient estimators \cite{ranganath2014black}, and the reparameterization trick \citep{kingma2013auto}. 

\subsection*{The score function gradient estimator }
The score function gradient estimator (also called Reinforce) is $\hat{g} := f(\mathbf{z})\nabla \log p(\mathbf{z};\theta)$. Its unbiasedness follows from the following computation, assuming the conditions of the dominated convergence theorem holds for $f$:
\begin{align*}
    \nabla_\theta \mathbb{E}_{p(\mathbf{z};\theta)}[f(\mathbf{z})] &= \int f(\mathbf{z}) \nabla_\theta p(\mathbf{z};\theta) d\mu(\mathbf{z})\\
    &= \int f(\mathbf{z}) \nabla_\theta\big(\log p(\mathbf{z};\theta)\big) p(\mathbf{z};\theta)  d\mu(\mathbf{z})\\
    &= \mathbb{E}\Big\{f(\mathbf{z})\nabla_\theta\log p(\mathbf{z};\theta)\Big\}.
\end{align*}
The estimator is generally applicable but in many cases has too high variance to be useful in practice. However, this estimator has proven useful in many situations with the inclusion of variance reduction techniques such as control variates \cite{ranganath2014black, tucker2017rebar, grathwohl2017backpropagation}. ~\looseness=-1
\begin{figure*}[!ht]
         \centering
         \includegraphics[scale = 0.55]{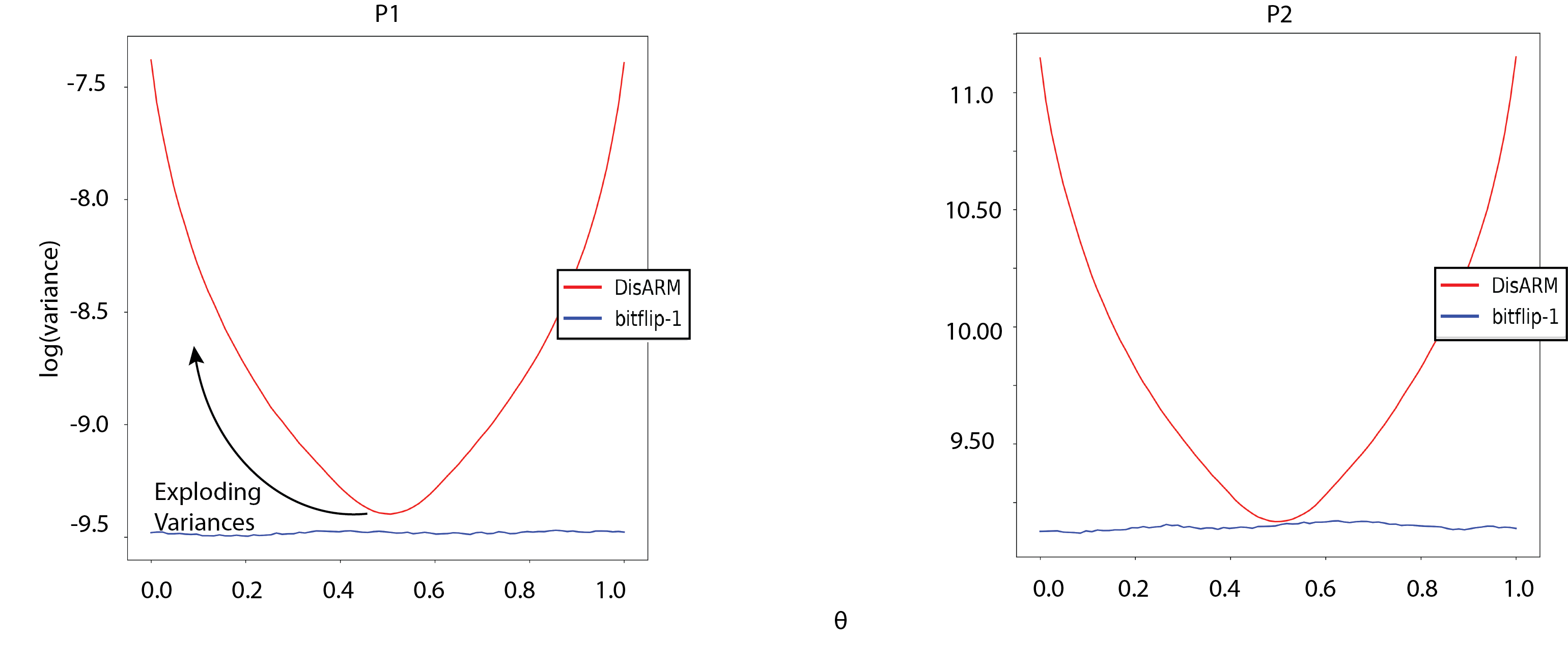}
     \caption{DisARM gradient variances potentially explode at the boundary of parameter space. \textit{Left:} variance curves for P1, $f(z) = \sum_{i=1}^K (z_i -t)^2$ \textit{Right:} variance curves for P2 $f(z) - (\sum_{i=1}^Kz_i -t)^2$. In both cases $K=20$, with $\theta_1 = \dots \theta_{19} = 0.5$ and $\theta_{20}$ varying on the x-axis.}
     \label{fig:illustrate}
\end{figure*}

\section*{The ARM and DisARM gradient estimators}
ARM \cite{yin2018arm} and DisARM (also called U2G) \cite{dong2020disarm,yin2020probabilistic} are two methods for reducing the variance of the score function gradient estimator estimator for Bernoulli latent variables. As notation, $\alpha_\theta$ will refer to the logits of the Bernoulli parameter, i.e. $\alpha_\theta := \log \frac{\theta}{1-\theta}$. The ARM estimator is motivated by a reparameterization. In one dimension, letting $b \sim \text{Logistic}(\alpha_\theta, 1)$ and $z = \mathbf{1}_{b > 0}$; the desired gradient $\nabla_\theta \mathbb{E}\big[f(z)\big] = \nabla_\theta \mathbb{E}_{b}\big[f(\mathbf{1}_{b>0})\big] = \mathbb{E}_{b}\big[ f(\mathbf{1}_{b>0})\nabla_\theta \log q_\theta(b)  \big]$ where $q_\theta$ is the likelihood of the Logistic distribution with parameter $\alpha_\theta$. Logistic random variables with identical marginal distributions can be sampled by letting $\epsilon \sim \text{Logistic}(0,1)$ and setting $b = \epsilon + \alpha_\theta$ and $\tilde{b} = -\epsilon  + \alpha_\theta$. This antithetic sampling produces an estimator with reduced variance:
\begin{align*}
    \hat{g}_{\text{ARM}} &:= \frac{1}{2}\big(f(\mathbf{1}_{b > 0})\nabla_\theta \log q_\theta(b) + f(\mathbf{1}_{\tilde{b} > 0})\nabla_\theta \log q_\theta(\tilde{b})\big)\\
    &= \frac{1}{2}\big(f(\mathbf{1}_{b > 0}) - f(\mathbf{1}_{\tilde{b} > 0}))\nabla_\theta \log q_\theta(b) \\
    &= \big(f(z) - f(\tilde{z})\big)(u - \frac{1}{2})\nabla_\theta\alpha_\theta.
\end{align*}
Here, $\sigma(\cdot)$ is the sigmoid operation and, $u$ is a uniform random variable defined by $\sigma(b - \alpha_\theta)$, and $z=\mathbf{1}_{1 -u < \theta}$, $\tilde{z}=\mathbf{1}_{u < \theta}$. The procedure naturally extends to the multi-dimensional case giving the estimator $\hat{g}_{\text{ARM}} = \big((f(\mathbf{z}) - f(\mathbf{\tilde{z}}))(\mathbf{u} - \frac{1}{2})\big)\nabla_\theta\alpha_\theta $\\

The DisARM estimator takes a conditional expectation of the ARM estimator, conditioning on the values $(z,\tilde{z})$:
\begin{align*}
    \hat{g}_{\text{DisARM}} &= \mathbb{E}_{p(b|z,\tilde{z})}\big[\hat{g}_{ARM}\big] \\
    &= \frac{1}{2}\big(f(z) - f(\tilde{z})\big)(-1)^{\tilde{z}}\mathbf{1}_{z\neq \tilde{z}} \sigma(|\alpha_\theta|)\nabla_\theta\alpha_\theta 
\end{align*}
This extends to the multi-dimensional case in an analogous way, requiring a constant number of function evaluations, and also further reduces  the variance of ARM estimator by nature of Rao-Blackwellization.

\section*{Variance properties of DisARM at the boundary}
Though DisARM is competitive compared to existing methods of gradient estimation, it has unfavorable variance at the boundaries of the parameter space. Reparameterizing the DisARM estimator in terms of probability $\theta$ gives:
\begin{align*}
    \hat{g}_{DisARM, j} = \frac{1}{2}\big(f(\mathbf{z}) - f(\mathbf{\tilde{z}})\big)\frac{1}{\min(\theta_j,1-\theta_j)}\mathbf{1}_{z_j \neq \tilde{z}_j} (-1)^{\tilde{z}_j}
\end{align*}
where $\tilde{\mathbf{z}}_j$ satisfies $\mathbb{P}[z_j = 0, \tilde{z}_j = 1] = \mathbb{P}[z_j = 1, \tilde{z}_j = 0] = \min(\theta_j,1-\theta_j)$, and $\mathbb{P}[z_j = \tilde{z}_j] = |1-2\theta_j|$.  

We analyze the variance as the difference $\mathbb{E}[(\hat{g}_{DisARM, j})^2]-\mathbb{E}[(\hat{g}_{DisARM, j})]^2$. Without loss of generality, considering the case where $\theta_j < \frac{1}{2}$, the expected square $\mathbb{E}[(\hat{g}_{DisARM, j})^2]$ is:
\vspace{2mm}
\begin{align}
    \resizebox{0.425\textwidth}{!}{$\mathbb{E}\Big[\frac{1}{4}\big(f(\mathbf{z}) - f(\tilde{\mathbf{z}})\big)^2 \frac{1}{\theta_j^2} \mathbf{1}_{z_j \neq \tilde{z}_j}\Big] = \frac{1}{2\theta_j}\mathbb{E}\Big[\big(f(\mathbf{z}_1^{(j)}) - f(\tilde{\mathbf{z}}_0^{(j)})\big)^2 \Big]$}  
    \label{eq:gsquare}
\end{align}
\vspace{2mm}where $\mathbf{z}_1^{(j)}$ and $\mathbf{\tilde{z}}_0^{(j)}$ are defined by hard-coding the $j$'th element of $\mathbf{z}$ as $1$ and $0$ respectively and sampling remaining shared elements from their respective distributions.  
For unbiased gradient estimators, the term $\mathbb{E}[\hat{g}]^2 = \big(\mathbb{E}\big[f(\mathbf{z})| z_j = 1\big] - \mathbb{E}\big[f(\mathbf{z})| z_j = 0\big]\big)^2$ are the same. Therefore, \Cref{eq:gsquare} suggests that DisARM suffers from large variances when $\theta_j \approx 1$ or $\theta_j \approx 0$ (see \Cref{fig:illustrate}). Another estimator competitive with DisARM is Reinforce-loo \cite{kool2019buy}, expressed as $\frac{1}{2\theta_j (1-\theta_j)}\Big((f(z_{1,j}) - f(z_{2,j}))(z_{1,j} - \theta_j) + (f(z_{2,j}) - f(z_{1,j}))(z_{2,j}- \theta_j) \Big)$ where now $\mathbf{z}_1$ and $\mathbf{z}_2$ are sampled independently. Again, the presence of the $\frac{1}{2\theta_j (1-\theta_j)}$ weight induces high variances at the boundary. This motivates us to consider estimators with bounded variance at the boundary. However, we note that this problem might be ameliorated by parameterizing $\theta$ by logits $\theta = \frac{e^\phi}{1+e^\phi}$ with $\nabla_\phi \theta = \theta(1-\theta)$ as is commonly done in practice. Though this parameterization avoids explicit enforcement of the [0,1] constraint during optimization, solutions at the boundary cannot be reached exactly. In our simulations, we have observed slower convergence of this approach relative to projected gradient descent in a number of problem settings.

\section*{Unbiased Monte Carlo estimate of the gradient via bit flips}
Note that the exact gradient is given by $E[f(\mathbf{z})|z_j = 1] - E[f(\mathbf{z})|z_j = 0]$. This suggests a simple estimation scheme: sample $\mathbf{z}\sim p_\theta$, and let $\tilde{\mathbf{z}}^{(j)}$ be the vector where the $j'th$ element of $\mathbf{z}$ is flipped. The single sample estimate is then $(-1)^{z_j}(f(\tilde{\mathbf{z}}^{(j)}) - f(\mathbf{z}))$.  We can apply this to all elements of the gradient for a single sample $\mathbf{z}$ and retain the unbiasedness property. Since this requires $O(K)$ function evaluations with K as the dimension of variable $\mathbf{z}$, which may be too expensive in many settings, we define and analyze \textit{bitflip}-1 as the randomized estimator given by sampling $\mathbf{z} \sim p_\theta$, sampling a random coordinate $j \sim \text{Categorical}(1,\cdots,K)$, and setting the estimate $\hat{g}_{\text{bitflip-1}, j} := K*(-1)^{z_j}(f(\tilde{\mathbf{z}}^{(j)})- f(\mathbf{z}))$, $\hat{g}_{\text{bitflip-1},-j} := 0$.  Interestingly, the only dependence of $\hat{g}$ on $\theta$ is through the sampling procedure. We also point out that though the DisARM estimator is shown to be uniformly minimum variance among estimators that employ linear combinations of antithetic sampled Bernoulli variables (\cite{yin2020probabilistic}, Proposition 2), bitflip-1 cannot be expressed in this manner (and moreover, the coordinates are no longer independent) and so is not dominated. In fact, bitflip-1 is lower variance than DisARM whenever $\frac{1}{2\min(\theta,1-\theta)} > K$. 

\begin{figure*}[!ht]
     \centering
     \includegraphics[width=\textwidth]{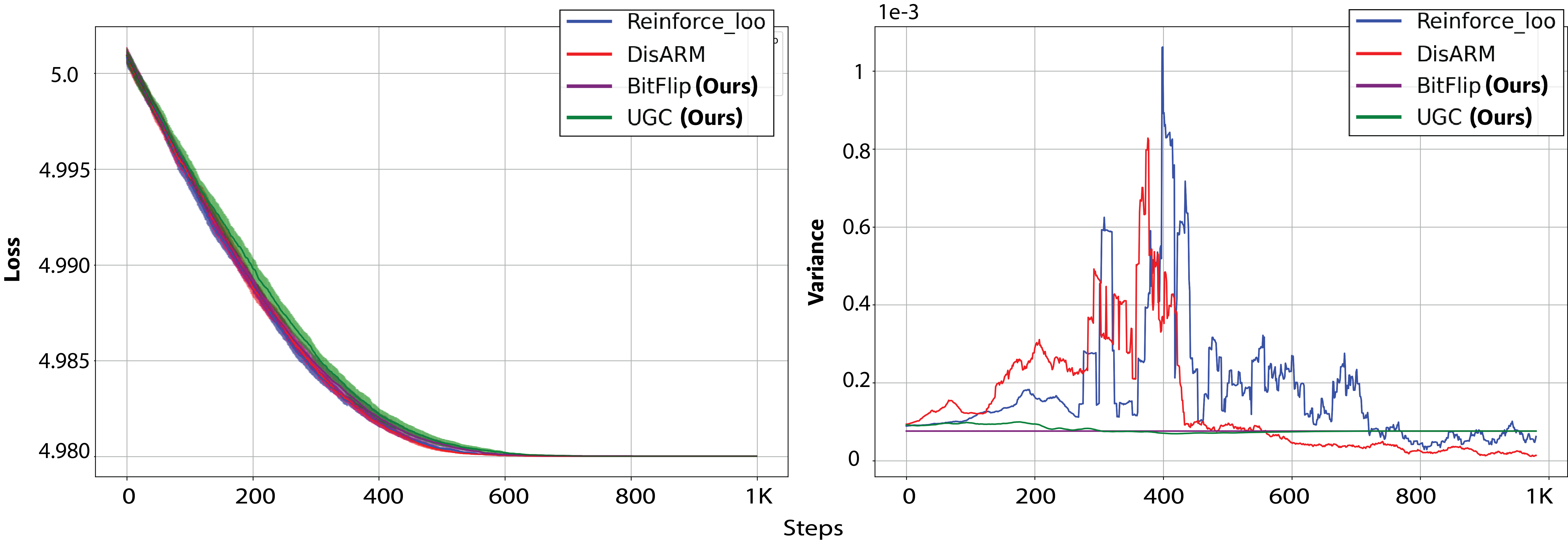}
     \caption{Performance on (P1) optimization problem $\min $ with $t = 0.499$, $K=20$. All methods tested converge to the optimal solution in this setting, though DisARM and Reinforce-loo suffer higher gradient variances. \textit{Left}: Training loss curves averaged over $10$ trials for the (P1) optimization problem with error bars $\pm \sigma/\sqrt{10}$ \textit{Right}: Gradient variances on (P1) averaged over $10$ trials.}
     \label{fig:2}
\end{figure*}

The expression of the gradient also suggests an interpretation of the DisARM estimator: that is, DisARM estimates $\mathbb{E}\big[f(\mathbf{z})| z_j = 1\big] - \mathbb{E}\big[f(\mathbf{z})| z_j = 0\big]$ with two samples and a multiplicative weight that ensures the unbiasedness property. Each of the two samples has $j$'th coordinate that is marginally $\text{Bernoulli}(\theta_j)$, with a joint distribution between the two samples that gives us maximal amount of information about the gradient. If we are limited to two function evaluations, this suggests considering estimates of the form $f(\mathbf{z}) - f(\mathbf{\tilde{z}})$ for some $(\mathbf{z}, \mathbf{\tilde{z}})$ with the marginal distribution $z_j \sim \text{Bernoulli}(\theta_j)$. However, with just two function evaluations it makes sense to disregard terms where $z_j = \tilde{z}_j$ as it is not clear how to construct an estimator of $\mathbb{E}[f(\mathbf{z})|z_j = 1] -  \mathbb{E}[f(\mathbf{z})|z_j = 0]$ in these cases. 

All of this suggests considering estimators of the form:

\begin{align}
    &\hat{g}_j := (-1)^{\tilde{z}_j}[f(\mathbf{z}) - f(\mathbf{\tilde{z}})] \\
    &\times \frac{1}{p[z_j = 1,\tilde{z}_j = 0] + p[z_j = 0,\tilde{z}_j = 1]}\mathbf{1}_{z_j \neq \tilde{z}_j}
\end{align}
where the correction term (Eq. 4) is to retain unbiasedness. This recovers DisARM when $p[z_j = 1,\tilde{z}_j = 0] = p[z_j = 0,\tilde{z}_j = 1] = \min(\theta, 1- \theta)$ and Reinforce-loo when $\mathbf{z}\indep \mathbf{\tilde{z}}$. An important fact about DisARM is that it maximizes $p[z_j = 0,\tilde{z}_j = 1] + p[z_j = 1,\tilde{z}_j = 0]$, i.e. the coupling given by $p[z_j = 0,\tilde{z}_j = 1] = \min(\theta_j,1-\theta_j)$ has the highest probability of differing values between $\tilde{z}_j$ and $z_j$ subject to the marginal constraint that each random variable is Bernoulli$(\theta_j)$. This is due to the fact that $p[z_j = 0,\tilde{z}_j = 1] \leq \theta$ and $p[z_j = 0,\tilde{z}_j = 1] \leq 1 - \theta$ following the two constraints given by $p[z_j = 0] = p[z_j = 0,\tilde{z}_j = 1] +  p[z_j = 0,\tilde{z}_j = 0]$ and $p[\tilde{z}_j = 1] = p[z_j = 1,\tilde{z}_j = 1] +  p[z_j = 0,\tilde{z}_j = 1]$. \\

However, it is clear that the minimum variance coupling depends on $f$ as we have:

\begin{align}
&E[\hat{g}^2] = \Big(\frac{1}{p[z_j = 0,\tilde{z}_j = 1] + p[z_j = 1,\tilde{z}_j = 0]}\Big) \\ &\times \mathbb{E}\big((f(\mathbf{z}) - f(\mathbf{\tilde{z}}))^2| z_j = 1, \tilde{z}_j = 0\big). 
\end{align}

\begin{figure*}[!ht]
     \centering
     \includegraphics[width=\textwidth]{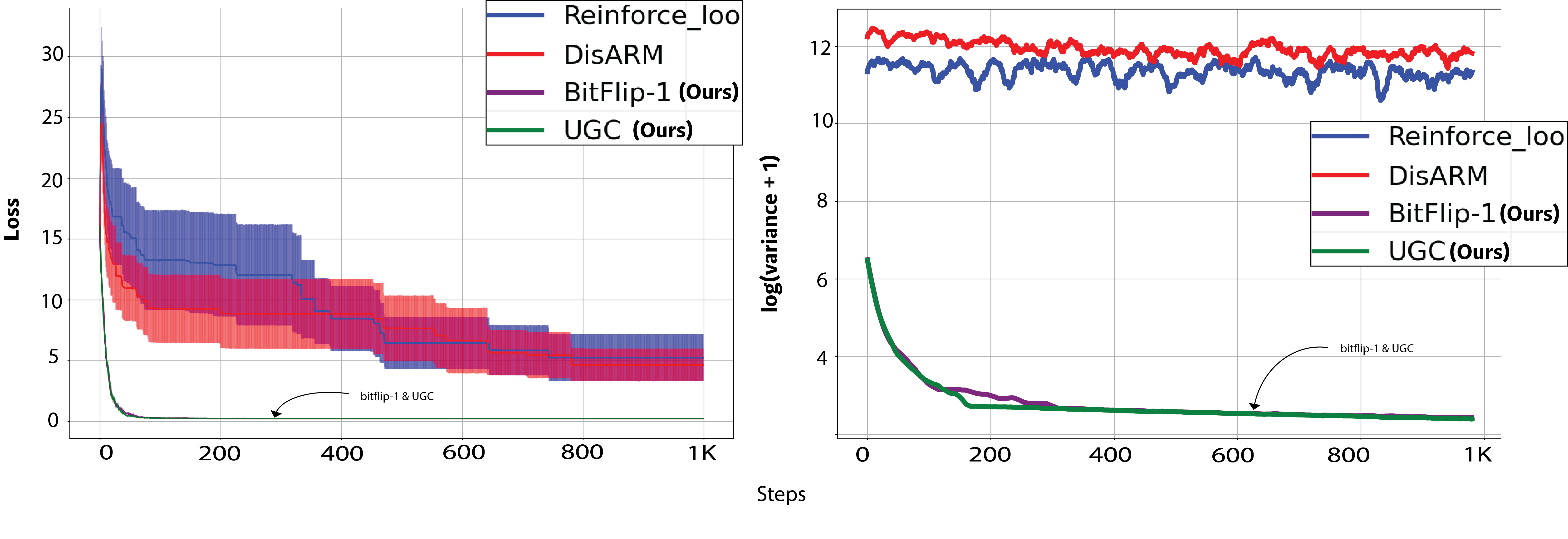}
     \caption{Performance on (P2) optimization problem  with $t = 0.499, K = 20$. DisARM and Reinforce-loo frequently fail to converge, while experiencing high gradient variances. \textit{Left}: Training loss curves averaged over $10$ trials for the (P2) optimization problem with errors bars $\pm \sigma/\sqrt{10}$ \textit{Right}: Average gradient variances on (P1) over 10 trials}
     \label{fig:3}
\end{figure*}

When $f$ is continuous (in the sense that $|f(\mathbf{z})-f(\tilde{\mathbf{z}})|$ is related to $d(\mathbf{z},\mathbf{\tilde{z}})$ for a distance metric $d$) there is a tradeoff between minimizing the first (Eq. 5) and second (Eq. 6) terms. As $p[z_j = 0,\tilde{z}_j = 1]$ (and $p[z_j = 1,\tilde{z}_j = 0]$) increase, the expected function differences (Eq. 6) are likely to be large. 
If $f$ is such that term (Eq. 6) tends to be large, independently sampled $\mathbf{z}$ and $\mathbf{\tilde{z}}$ may even be lower variance than antithetic samples \cite{dong2020disarm}. DisARM updates the largest number of terms possible by maximizing the probabilities $p[z_j = 0,\tilde{z}_j = 1]$ and $p[z_j = 1,\tilde{z}_j = 0]$ and hence minimizes term (Eq. 5), but insodoing may incur high variance through large values of term (Eq. 6).

\section*{Variance properties of \textit{bitflip}-1}
Without loss of generality assume $\theta_i < 0.5$ and consider the variance of a single coordinate of each estimator. The argument can be easily extended to $\theta \geq 0.5$. We also assume the following natural continuity property of the function $f$: 
\begin{assumption} Given four binary vectors $z,w,\tilde{z},\tilde{w} \in \{0,1\}^K$, 
if $\{j: \tilde{z}_j \neq z_j\} \supset \{j: \tilde{w}_j \neq w_j\}$ and $w_i = z_i$ for all $i$ such that $w_i = \tilde{w}_i$ and $z_i = \tilde{z}_j$, then $|f(w) - f(\tilde{w})| \leq  |f(z) - f(\tilde{z})|$. 
\end{assumption}

In other words, given two binary strings we cannot make their function evaluations closer by introducing additional coordinates where they differ. Since each estimator considered is unbiased, it suffices to consider $\mathbb{E}[\hat{g}^2]$ for each gradient estimator $\hat{g}$.
\begin{proposition} (Variance of bitflip-1) Let $\hat{g}$ be an estimator in the family of estimators given by (Eq. 3-4), which includes DisARM and Reinforce-loo gradient estimators. If Assumption 1 holds and if $\frac{1}{2\min(\theta_j, 1-\theta_j)} \geq K$:
\begin{align*}
    \text{Var}(\hat{g}_{\text{bitflip-1}}) \leq \text{Var}(\hat{g}) 
\end{align*}
\end{proposition}
We present an expanded version of this proposition and proof in the appendix. We also note that when $f(z)$ is separable, bitflip-1 has uniformly lower variance than DisARM in the following sense:
\begin{proposition} Consider a member of the family of estimators given by (Eq. 3-4), which includes DisARM and Reinforce-loo and denote this estimator $\hat{g}$. If $f(\mathbf{z}) = \sum_{i=1}^K h(z_i)$, then:
\begin{align*}
    \min_{\theta_1,\dots, \theta_K} \max_{j=1,\dots,K} Var(\hat{g}_{j})  \geq  \max_{\theta_1,\dots, \theta_K} \max_{j=1,\dots,K}Var(\hat{g}_{\text{bitflip}, j})
\end{align*}
\end{proposition}

\section*{Unbiased Gradient Variance Clipping}
Though bitflip-1 has bounded variance for a given latent variable dimension $K$, it's variance grows linearly with $K$. Meanwhile, DisARM has variance growing with $\frac{1}{\min(\theta, 1-\theta)}$ despite only depending on $K$ implicitly through the function $f$. Motivated by these complementary behaviors and fact that $\theta_j$ and $K$ are available, we can construct an estimator that dominates DisARM as follows.
\begin{equation}
  \hat{g}_{UGC,j} =
    \begin{cases}
      \hat{g}_{bitflip-1,j} & \text{if} \min(\theta_j, 1-\theta_j) < \tau\\
      \hat{g}_{DisARM,j} & \text{if} \min(\theta_j, 1-\theta_j) \geq \tau\\
    \end{cases}         
\end{equation}
where $\tau$ is a tuning parameter of the estimator. We denote this estimator by \textit{unbiased gradient variance clipping} (UGC) as it replaces potentially high variance gradient estimates with bounded variance estimates without breaking unbiasedness of the estimate. A standard choice of $\tau$ is $\frac{1}{2K}$, motivated by the following result:

\begin{proposition}(Variance of UGC) Under assumption 1, when $\tau \leq \frac{1}{2K}$, $Var(\hat{g}^{(UGC)}) \leq Var(\hat{g})$ for any $\hat{g}$ in the family of estimators given by (Eq. 3-4), which includes DisARM and Reinforce-loo gradients.

\end{proposition}


We find that UGC achieves better performance than bitflip$-1$ and DisARM on a number of tasks.

\section*{Experiments}
\begin{figure*}[!ht]
         \centering
         \includegraphics[width=\textwidth]{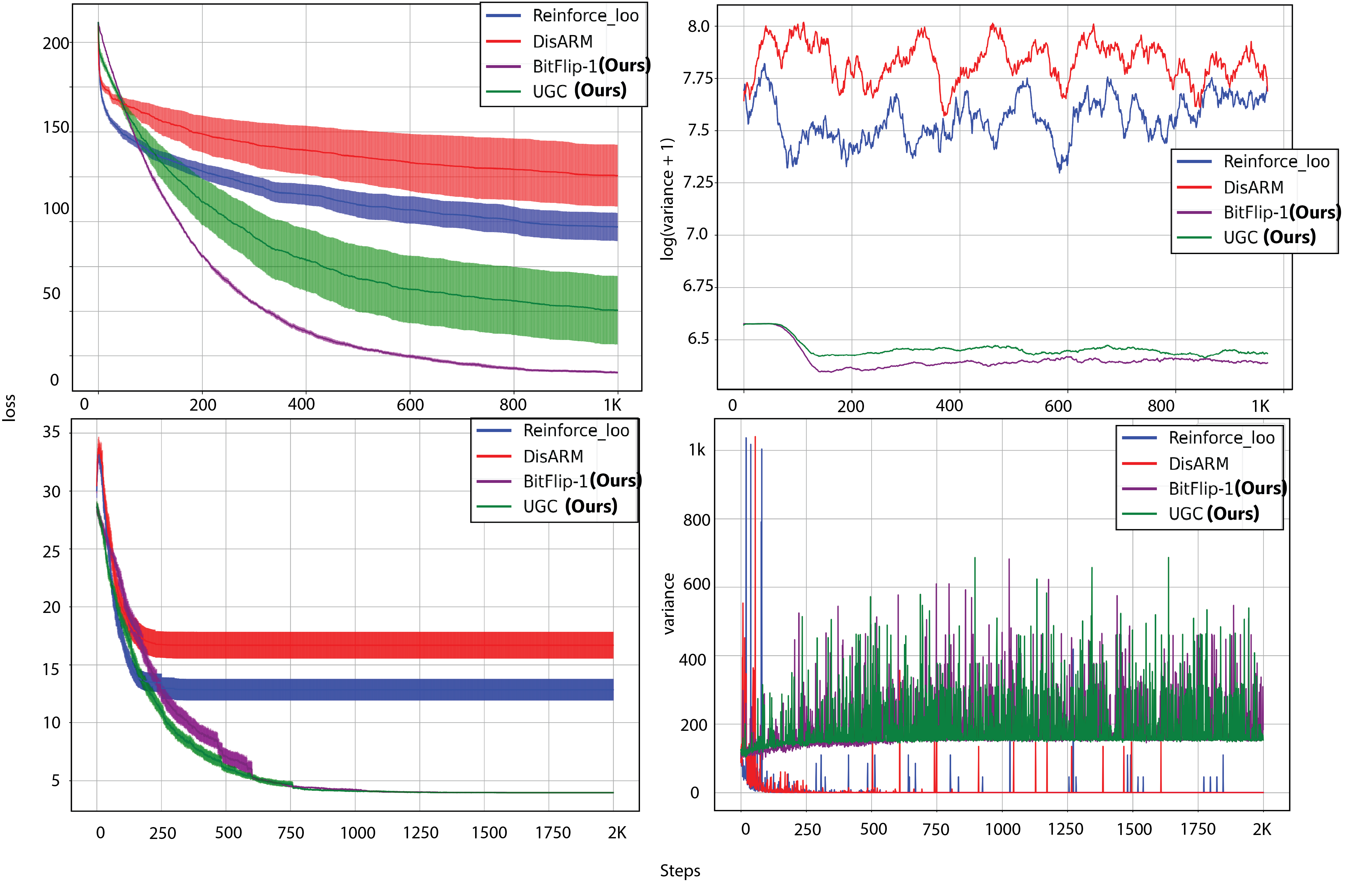}
     \caption{Performance on the gradient based subset optimization problem for linear regression. $p = 200$, $n=60$, $\Sigma = I$, $|S| = 3$,  \textit{top}: SNR $= \beta^\top \beta/\sigma^2 = 3.8125$, parameterization by $\phi = \log(\theta/(1-\theta))$\textit{bottom}: SNR $= \beta^\top \beta/\sigma^2 =1.694$. Parameterization by $\theta$, with projected gradient descent onto $[0,1]$. \textit{Left}: Training loss curves for the best subset optimization problem, averaged over $10$ random samples of the data with error bars $\pm \sigma/\sqrt{10}$. \textit{Right}: Average gradient variances across $10$ random samples of the data. Though bitflip-1 and UGC are higher variance in the second example, we note that this is because they are in the correct part of parameter space}
\end{figure*}

\subsection*{Toy experiments}
In \cite{tucker2017rebar}, the authors optimize the objective $\mathbb{E}_\theta[(z-t)^2]$ where $z$ is a single Bernoulli random variable with a parameter $\theta$ and $t$ is set to either $0.49$ or $0.499$. The optimizer of this problem is $\theta = 0$, with values of $t$ closer to $0.5$ representing harder problems. As bitflip-1 computes the exact gradient for univariate latent variable $z$, we extend this problem to two multivariate problems:
\begin{align*}
   \resizebox{0.48\textwidth}{!}{ $(P1):\min \mathbb{E}[\sum_{k=1}^K (z_k - t)^2];~~(P2):\min \mathbb{E}[(\sum_{k=1}^K z_k - t)^2]$}
\end{align*}
In problem (P1), due to the separability of the objective, bitflip-1 computes the exact gradient multiplied by $K$ and updates a random component (\Cref{fig:2}). Problem (P2) is harder in the sense that it contains many interaction terms and the exact gradient is expensive to compute for moderate $K$. \Cref{fig:3} shows results for $K=20$ and $t = 0.499$ (with other results in the appendix). Notably, for $(P2)$ both the Reinforce-loo baseline and DisARM fail to converge to the optimum. This occurs due to the fact that these gradients can often be in the wrong direction due to noise and then are unable to estimate high magnitude gradients at $\theta =1$. When $\theta \approx 1$, UGC will switch to using bitflip gradients and can move away from the suboptimal $\theta = 1$. 
\begin{figure*}[!ht]
         \centering
         \includegraphics[width=\textwidth]{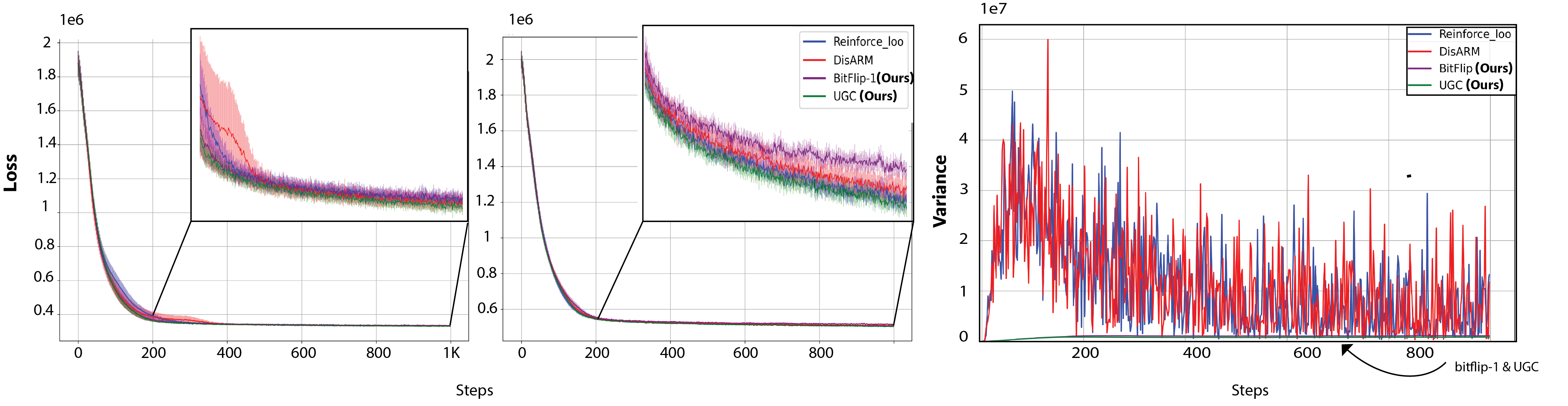}
     \caption{Performance on the gaussian mixture model problem fit via discrete variational autoencoders. Cluster means are sampled $N(0, 8^2)$ per simulation. \textit{Right}: Training loss curves for the gaussian mixture model problem ($\sigma = 2.0$), averaged over $10$ random samples of the data with error bars $\pm \sigma/\sqrt{10}$. \textit{Middle}:Training loss curves for $\sigma = 4.0$ \textit{Right}: Average gradient variances ($\sigma = 4.0$) across $10$ random samples of the data. Through the experiment, the true number of clusters is $6$, the number of features is $20$, and the hidden dimension is $10$}
\end{figure*}
\begin{figure*}[!ht]
         \centering
         \includegraphics[width=\textwidth]{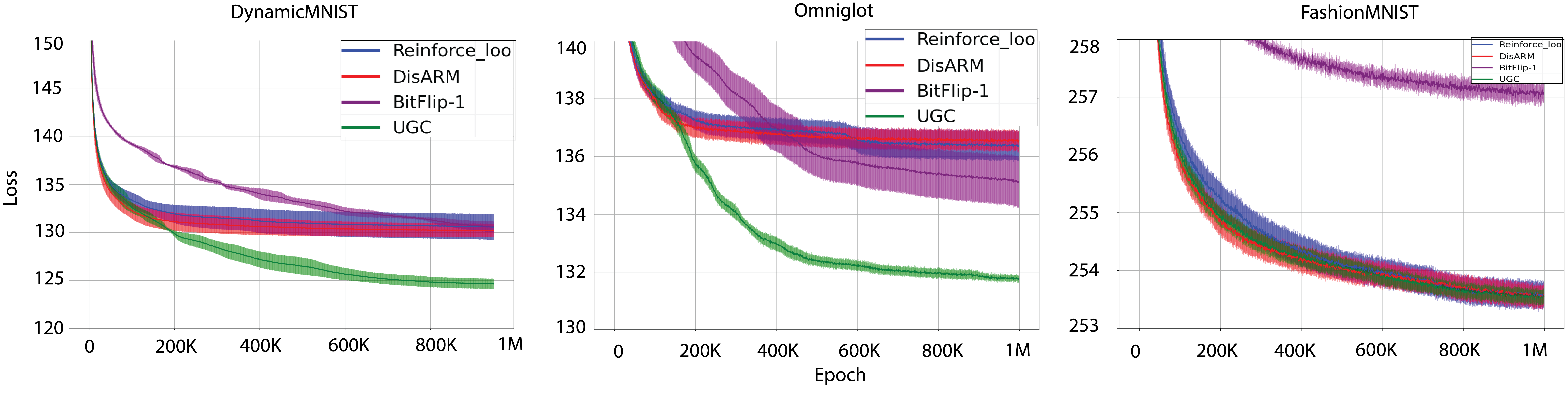}
     \caption{Performance on the binarized discrete VAE fit to DynamicMNIST, FashionMNIST and Omniglot datasets over 5 random seeds, with error bars given by $\pm \sigma/\sqrt{5}$. The binary latent variable is 30 dimensional with 1-layer encoder and decoder networks. UGC achieves better convergence than alternative estimators.}
\end{figure*}
\subsection*{$L_0$ best subset regression} Fitting linear regression with a sparsity penalty has become a ubiquitous task across many domains \cite{tibshirani2011regression}. Such regression estimators frequently are computed by minimizing squared error subject to a constrain on the $L_1$ norm of the regression coefficients $\beta$. The non-convex problem of optimizing subject to constraint on the $L_0$ norm has received less attention due to computational challenges but is addressed in \cite{yin2020probabilistic}. Specifically, they consider the following estimator of $\beta$ under the linear regression assumptions $y\sim \mathcal{N}(x^\top \beta, \sigma^2)$:
\begin{align*}
    \min_{\beta}\frac{1}{n} ||\mathbf{y} - \mathbf{X}\beta||_2^2 + \lambda ||\beta||_0
\end{align*}

This optimization problem penalizes the cardinality of the coefficient vector $\beta$, rather than its $L_1$ norm and so more directly encodes the assumption that the true coefficient vector is sparse.  In \cite{yin2020probabilistic}, the authors show that this problem can be approximately solved with the gradient estimator DisARM via the equivalent optimization problem:  $\min_{\theta}\mathbb{E}_{z \sim \theta}\big[\min_{\beta} \frac{1}{n} || \mathbf{y} - \mathbf{X}(\mathbf{z} \odot \beta)||_2^2 + \lambda || \mathbf{z}||_0\big]$ where $\odot$ means elementwise multiplication. The solutions of the second problem are guaranteed to occur at the boundaries of the parameter space and coincide with the solution of the original regression problem. As the solutions occur at the boundary, this scenario is one where bitflip-1 and UGC perform well, shown in Figure 4. Specifically, in low signal-to-noise (SNR) settings, other gradient estimators cannot reliably recover the correct solution (Tables 1 and 2).  

\begin{table}
    \centering
    \caption{False Positive Rate (FPR) of best subset selection.}
    \rowcolors{5}{}{gray!10}
    \begin{tabular}{*5c}
        \toprule
        & \multicolumn{4}{c}{Gradient estimator} \\
        \cmidrule(lr){2-5}
        SNR & bitflip-1  & UGC & DisARM & Rein.-loo\\   
        \midrule
        15.25 &     \textbf{0.0 (0.0)}  & \textbf{0.0 (0.0)} & 0.05 (0.02) & 0.04 (0.01)  \\
        3.81 &  \textbf{0.0 (0.0)} & \textbf{0.0 (0.0)}   & 0.06 (0.03) & 0.04 (0.01)\\
        1.69 &   \textbf{0.01 (0.01)}  & \textbf{0.01 (0.01)} & 0.06 (0.03) & 0.05 (0.02) \\
        0.95 & 0.04 (0.01) &    \textbf{0.03 (0.01)}   &    0.06 (0.02) & 0.05 (0.01)    \\
        \bottomrule
    \end{tabular}
\end{table}

\begin{table}
    \centering
    \caption{True Positive Rate (TPR) of best subset selection.}
    \rowcolors{5}{}{gray!10}
    \begin{tabular}{*5c}
        \toprule
        & \multicolumn{4}{c}{Gradient estimator} \\
        \cmidrule(lr){2-5}
        SNR & bitflip-1  & UGC & DisARM & Rein.-loo\\   
        \midrule
        15.25 &     \textbf{0.96 (0.1)}  & \textbf{0.96 (0.1)} & 0.56 (0.26) & 0.6 (0.36)  \\
        3.81 &  \textbf{1.0 (0.0)} & \textbf{1.0 (0.0)}   & 0.66 (0.26) & 0.53 (0.16)\\
        1.69 &   0.83 (0.27) & \textbf{0.87 (0.22)} & 0.43 (0.26) & 0.50 (0.31) \\
        0.95 & 0.43 (0.30) &    \textbf{0.67 (0.21)}   &    0.40 (0.29) & 0.43 (0.21)  \\
        \bottomrule
    \end{tabular}
\end{table}

\subsection*{Gaussian mixture model} We investigate the capability of a discrete variational autoencoder fit with each gradient estimator to identify Gaussian mixtures. Specifically we generate samples from a $20-$dimensional Gaussian mixture model distribution with $6$ components by first sampling component means from a $N(0,8^2)$ distribution, then sampling data conditional on component means from a Normal distribution with variance $\sigma^2$, with $\sigma^2$ being the parameter controlling the signal to noise ratio. Though each esimator achieves comparable convergence rate for multiple signal to noise ratios, bitflip-$1$ and UGC have markedly lower variance throughout training (Figure 5).

\subsection*{Discrete variational autoencoder training} We replicate the discrete variational autoencoder architecture and experimental setup on binarized DynamicMNIST, Omniglot and FashionMNIST datasets from \cite{yin2018arm} and \cite{dong2020disarm}. Interestingly, we note that DisARM exhibits fast convergence early on in training but later in training is unable to make progress, while bitflip-1 proceeds slowly during initial training but reaches a better final optimum. UGC achieves the best of both worlds: after switching to bitflip-1 derived gradients, it reaches a better solution than both methods (Figure 6).

\section*{Discussion} We have presented a method for producing low variance gradient estimates at the boundary of the parameter space for Bernoulli latent variable models. Noticing that existing methods suffer high variance gradients near the boundary of $[0,1]$, we introduce a combined estimator, UGC, that uses DisARM gradients near the middle of $[0,1]$ and bitflip-1 gradients near the boundary. We expect our approach to be useful for fitting various kinds of sparse latent variable models; for example, for fitting variational autoencoders with spike and slab priors via mean field variational inference \cite{moran2021identifiable}. Our empirical results hopefully open the door to a number of theoretical questions. Future work may define classes of discrete functions and estimators where we can find optimal gradient estimators subject to constraint on the number of function evaluations. 

\newpage 
\smallskip \noindent

\bibliography{refs} 
\onecolumn
 \newpage
\section*{Supplementary Material}
\subsection*{Further background: Reparameterization trick} Another gradient estimator is given by the reparameterization trick \citep{kingma2013auto}, which requires $f(\mathbf{z})$ to be differentiable, and for $\mathbf{z}$ to be expressable as a differentiable transformation of exogenous noise $\mathbf{z} = T(\theta, \epsilon)$, where $\epsilon \sim g(\cdot)$ is free of $\theta$. When this holds, an unbiased estimator of $\nabla \mathbb{E}_{p(\mathbf{z};\theta)}[f(\mathbf{z})]$ is $\nabla_{T}f(T(\epsilon, \theta))\big(\frac{\partial}{\partial \theta} T(\theta,\epsilon)\big)$, where the second term is the Jacobian matrix of the transformation. Unbiasedness follow from a change of variables: $\mathbb{E}_{p(\mathbf{z};\theta)}[f(\mathbf{z})] = \mathbb{E}_{\epsilon}\big[f(T(\theta,\epsilon))\big]$, and then applying the chain rule. The reparameterization gradient estimator is lower variance than score function gradient estimator, but less generally applicable \citep{naesseth2017reparameterization}. In the context of discrete random variables, it's necessary to apply a continuous relaxation to $\mathbf{z}$ and extend the domain of $f$ to account for continuous input.
\subsection*{Exact gradient}
The expression of the exact gradient as $\mathbb{E}\big[f(\mathbf{z})| z_j = 1\big] - \mathbb{E}\big[f(\mathbf{z})| z_j = 0\big]$ is seen as follows:
\begin{align*}
    \nabla_{\theta_j} \mathbb{E}f(z_1,\dots, z_k) &= \sum_{\mathbf{z} \in \{0,1\}^k} f(z_1,\dots, z_k) \times \nabla_{\theta_j} \prod_{i=1}^k \theta_i^{z_i}(1-\theta_i)^{1-z_i} \\
    &= \sum_{\mathbf{z}: z_j = 1} f(z_1,\dots,z_k) \prod_{i\neq j}\theta_i^{z_i}(1-\theta_i)^{1-z_i} - \sum_{\mathbf{z}: z_j = 0} f(z_1,\dots,z_k) \prod_{i\neq j}\theta_i^{z_i}(1-\theta_i)^{1-z_i}\\
    &= \mathbb{E}\big[f(\mathbf{z})| z_j = 1\big] - \mathbb{E}\big[f(\mathbf{z})| z_j = 0\big]
\end{align*}

\subsection*{Step up UGC procedure (tUGC)}
Though UGC lowers the variance of DisARM, it is not the optimal aggregation procedure. This is due to the fact that the procedure is still unbiased if only a subset of the coordinates can be chosen to be updated. If we are choosing only a subset of the coordinates it makes sense to choose the smallest values of $\min(\theta_j, 1-\theta_j)$ to be updated via bitflip-1 rather than DisARM, as these would have the highest variance DisARM gradients. Let $\tilde{\theta}_{(1)}, \dots, \tilde{\theta}_{(K)}$ be the sorted values of $\min(\theta_j, 1-\theta_j)$ and $\sigma(\cdot)$ the reverse permutation. Consider $\hat{T} := \sup\{T : \tilde{\theta}_{(T)} \leq \frac{1}{2T}\}$. Sample $q\sim \text{Categorical}(1,\dots, \hat{T})$ and update $\hat{g}_{\sigma(q)} = \hat{T}[f(\mathbf{z}_1^{\sigma(q)}) - f(\mathbf{z}_0^{\sigma(q)})]$ with other $\hat{g}_j$ corresponding to the $\hat{T}$ lowest values of $\tilde{\theta}$ set to $0$. The remaining indices corresponding to larger values of $\tilde{\theta}$ are set to the DisARM estimate.  This is lower variance than UGC as $\hat{T}\leq K$. We denote this modification tUGC. After a few gradient updates, tUGC tends to behave like bitflip-1 (shown in Figure 7). However, these first few steps can be quite important (Figure 8-9) as tUGC vastly outperforms bitflip-1 for VAE training. As performance is quite similar to UGC overall we report results from UGC in the main text. 

\begin{figure}
    \centering
    \includegraphics{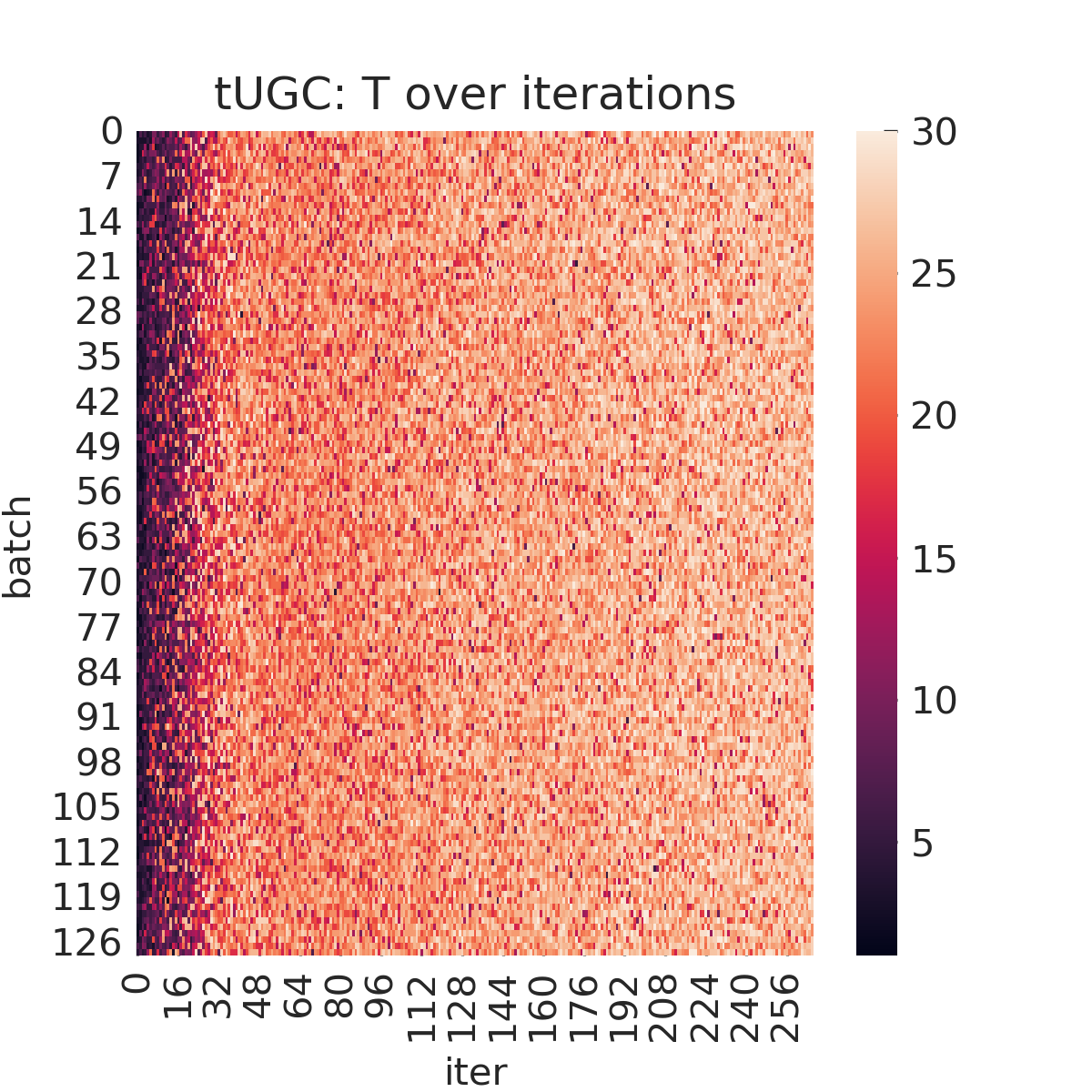}
    \caption{Convergence of tUGC to bitflip-1 over stochastic gradient descent iterations. Here, $K=30$ and the value of $\hat{T}$ approaches $K$ for each coordinate}
    \label{fig:my_label}
\end{figure}
\subsection*{Proof of Proposition 1} We present an expanded version of the proposition statement. First we define the $K-$sample versions of each of the estimators as follows. For bitflip, $\hat{g}_{\text{bitlfip-k}, j} = f(\mathbf{z}_1^{(j)}) - f(\mathbf{z}_0^{(j)})$ (requiring K+1 function evalutions) and  $\hat{g}_{\text{DisARM-k}, j} = \frac{1}{K} \sum_{k=1}^K \hat{g}_{\text{DisARM}, j}^{(k)} $ (with each $\hat{g}_{\text{DisARM}, j}^{(k)}$ being an independently generated instance of the DisARM estimator). The definition of $\hat{g}_{\text{Reinforce-loo-k}}$ is analogous. The latter two estimators require $2K$ function evaluations. 
\begin{proposition} Assume that $\hat{g}$ is an estimator of the gradient that can be expressed according to (Eq.3-4). and $\hat{g}_k := \frac{1}{K} \sum_{i=1}^K\hat{g}^{(i)}$ for independently generated $\hat{g}^{(i)}$. Under Assumption 1, we have:
\begin{itemize}
    \item $Var(\hat{g}_{\text{bitflip-k}}) \leq Var(\hat{g}) $
    \item $Var(\hat{g}_{\text{bitflip-k}}) \leq Var(\hat{g}_{\text{Reinforce}}) $; if additionally $f \geq 0$ or $f \leq 0$
\end{itemize}
If $\frac{1}{2\min(\theta_j, 1-\theta_j)} \geq K$:
\begin{itemize}
    \item $Var(\hat{g}_{\text{bitflip-1}}) \leq Var(\hat{g}) $
    \item $Var(\hat{g}_{\text{bitflip-k}}) \leq Var(\hat{g}_{k})$
    \item $Var(\hat{g}_{\text{bitflip-1}}) \leq Var(\hat{g}_{\text{Reinforce}}) $; if additionally $f \geq 0$ or $f \leq 0$
\end{itemize}
\end{proposition}

\textit{Proof.} The \textit{bitflip}-K estimator for $\nabla_{\theta_j}\mathbb{E}\big[f(\mathbf{z})\big]$ is:

$$\hat{g}_{\text{bitflip}-K,j} := f(\mathbf{z}_1^{(j)}) - f(\mathbf{z}_0^{(j)})$$
while the \textit{bitflip}-1 estimator is 
$$\hat{g}_{\text{bitflip}-1} := K*\big(f(\mathbf{z}_1^{(j)}) - f(\mathbf{z}_0^{(j)})\big)\mathbf{1}_{q = j}$$
where $\mathbf{z}\sim p_{\theta}$ is a sample from the given factorial Bernoulli distribution, $\mathbf{z}_1^{(j)}$ is $\mathbf{z}$ with it's $j$'th element set to $1$, $\mathbf{z}_0^{(j)}$ has its $j$'th element set to $0$, and $q\sim \text{Categorical}(1,\dots, K)$. For the first claim, we consider the bitflip K-sample estimator:
\begin{align*}
\mathbb{E}[\hat{g}_{j}^2] &= \frac{1}{ (\mathbb{P}[z_j = 0, \tilde{z}_j = 1] + \mathbb{P}[z_j = 1, \tilde{z}_j = 0])}\mathbb{E}\Big[\big(f(\mathbf{z}_1^{(j)}) - f(\tilde{\mathbf{z}}_0^{(j)})\big)^2 \Big] \\
&\geq \mathbb{E}\Big[\big(f(\mathbf{z}_1^{(j)}) - f(\tilde{\mathbf{z}}_0^{(j)})\big)^2 \Big]
\end{align*}
 Our assumed continuity condition allows us to conclude $\mathbb{E}[\hat{g}^2] \geq \mathbb{E}[\hat{g}_{\text{bitflip}-k}^2]$, and so has uniformly lower variance. 
For the second bullet, the score function gradient estimator is given by 
$$\hat{g}_{\text{R}, j} := f(\mathbf{z})\nabla_\theta \log p_{\theta}(\mathbf{z})  = f(\mathbf{z}) \Big[\frac{1}{\theta_j}z_j - \frac{1}{1-\theta_j}(1-z_j)\Big]$$
A calculation shows that the expected square is:
\begin{align*}
    \mathbb{E}[\hat{g}_{\text{R}, j}^2] &= \frac{\mathbb{E}[f(\mathbf{z}_1^{(j)})]^2}{\theta_j} + \frac{\mathbb{E}[f(\mathbf{z}_0^{(j)})]^2}{1 - \theta_j} \\
    &\geq \mathbb{E}[f(\mathbf{z}_1^{(j)})^2 + f(\mathbf{z}_0^{(j)})^2] \\
    &\geq \mathbb{E}[\big(f(\mathbf{z^{(j)}}_1) - f(\mathbf{z^{(j)}}_0)\big)^2]
\end{align*}
where the last line follows when $f\geq 0$ or $f \leq 0$. This shows the lower variance of bitflip-k when compared to Reinforce. We also see that the bound holds with: 
\begin{align*}
    \mathbb{E}[\hat{g}_{\text{R}, j}^2] &\geq \frac{1}{\min(\theta_j,1-\theta_j)}\mathbb{E}[\big(f(\mathbf{z}_1^{(j)}) - f(\mathbf{z}_0^{(j)})\big)^2]
\end{align*}
and hence, Reinforce is higher variance than bitflip-1 whenever $\min(\theta,1-\theta) < \frac{1}{K}$ which is implies when $\min(\theta,1-\theta) < \frac{1}{2K}$. (5th bullet point). The 3rd and 4th results come from 
\begin{align*}
\mathbb{E}[\hat{g}_{j}^2] &= \frac{1}{ (\mathbb{P}[z_j = 0, \tilde{z}_j = 1] + \mathbb{P}[z_j = 1, \tilde{z}_j = 0])}\mathbb{E}\Big[\big(f(\mathbf{z}_1^{(j)}) - f(\tilde{\mathbf{z}}_0^{(j)})\big)^2 \Big] \\
&\geq K\mathbb{E}\Big[\big(f(\mathbf{z}_1^{(j)}) - f(\tilde{\mathbf{z}}_0^{(j)})\big)^2 \Big] \\
&\geq K\mathbb{E}\Big[\big(f(\mathbf{z}_1^{(j)}) - f(\mathbf{z}_0^{(j)})\big)^2\Big] 
\end{align*}
where we use the fact that $\mathbb{P}[z_j = 0, \tilde{z}_j = 1] + \mathbb{P}[z_j = 1, \tilde{z}_j = 0] \leq 2\min(\theta_j, 1-\theta_j)$ and Assumption 1. Dividing by $K$ on both sides of the inequality leads to the result for the K-sample estimators.

\subsection*{Comparing bitflip-1 to the class of coordinate-wise independent estimators}
\begin{theorem} Consider a function $f(\mathbf{z}) = \sum_{i=1}^K h(z_i)$. Suppose that an estimator $\hat{g}$ is of the following form: 
\begin{align}
    \hat{g}_j = \frac{(-1)^{\tilde{z}_j}\mathbf{\mathbf{1}_{z_j \neq \tilde{z}_j}}}{p_1(\theta_j) + p_0(\theta_j)} \big[ f(\mathbf{z}) - f(\tilde{\mathbf{z}})\big]
\end{align}
where $p_1(\theta_j) := P[z_j = 1, \tilde{z}_j = 0]$ and  $p_0(\theta_j) := P[z_j = 0, \tilde{z}_j = 1]$ and marginally $z_j \sim \text{Bernoulli}(\theta_j)$, $\tilde{z}_j \sim \text{Bernoulli}(\theta_j)$, a general class of estimators which includes DisARM when $p_1(\theta_j) = p_0(\theta_j) = \min(\theta_j, 1-\theta_j)$ and Reinforce-loo when $p_1(\theta_j) = p_0(\theta_j) = \theta_j(1-\theta_j)$ . Then the following holds:
\begin{align*}
    \min_{\theta_1, \dots, \theta_K} \max_{j = 1,\dots, K} \text{var}(\hat{g}_j) \geq (K-1)(h(1) - h(0))^2 = var(\hat{g}_{\text{bitflip-1}}) 
\end{align*}
\end{theorem}
The proof requires the lower bound:
\begin{align*}
    \text{var}(\hat{g}) &\geq \mathbb{E}[\text{var}(\hat{g} | \mathbf{1}[z_j \neq \tilde{z}_j])] \\ 
    &= \mathbb{E}\Big[\mathbf{1}[z_j \neq \tilde{z}_j] \frac{1}{(p_1(\theta_j) + p_0(\theta_j) )^2}\sum_{i\neq j}\text{var}(h(z_i) - h(\tilde{z}_i))\Big] \\
    &=\mathbb{E}\Big[\mathbf{1}[z_j \neq \tilde{z}_j] \frac{1}{(p_1(\theta_j) + p_0(\theta_j) )^2}\sum_{i\neq j} (h(1)-h(0))^2 (p_1(\theta_i) + p_0(\theta_i)) \Big]\\
    &= \frac{(h(1)-h(0))^2}{p_1(\theta_j) + p_0(\theta_j)} \sum_{i\neq j} (p_1(\theta_i) + p_0(\theta_i)) 
\end{align*}
Now choose $j$ as the one corresponding to one of the smallest values of $p_1(\theta_j) + p_0(\theta_j)$. We have:
\begin{align*}
    \text{var}(\hat{g}) &\geq ((h(1)-h(0))^2\sum_{i\neq j} \frac{p_1(\theta_i) + p_0(\theta_i)}{p_1(\theta_j) + p_0(\theta_j)} \\
    & \geq ((h(1)-h(0))^2 \sum_{i\neq j} 1 \\
    & = (K-1) ((h(1)-h(0))^2
\end{align*}

This implies the result of Proposition $5$ as a special case of $h$ and $\hat{g}$. This class of estimators includes the Reinforce-loo estimator when $\mathbf{z}$ and $\mathbf{\tilde{z}}$  are independent and DisARM when $\mathbf{z}$ and $\mathbf{\tilde{z}}$ are antithetic. \\

When $h_i$ are allowed to have dependence on $i$ we have a weaker result, so long as each $h_i$ is injective. For simplicity of notation assume $p_0(\theta_j) = p_1(\theta_j) =: p(\theta_j)$:

\begin{align*}
    \frac{\text{var}(\hat{g}_j)}{\text{var}(\hat{g}_{\text{bitflip},j})} = \frac{1 - 2p(\theta_j)}{2(K-1)p(\theta_j)} + \frac{1}{K-1}\sum_{i\neq j} \frac{p(\theta_i)(h_i(1) - h_i(0))^2}{p(\theta_j)(h_j(1) - h_j(0))^2}
\end{align*}
Due to the second term there is at least one $j$ such that this variance ratio is at strictly greater than $1$ when all $\theta_j \neq 0.5$. Notice that when $2Kp(\theta_j) <1$ the variance ratio is greater than $1$.  
\subsection*{Additional Results for P1 and P2} We start by deriving exact variances for the bitflip-1 estimator for these problems. Let $q \sim \text{Categorical}(1,\dots,K)$ be the categorical random variable that selects a coordinate to update. The variance of the $j'$th coordinate of the gradient estimate for P1 is then:

\begin{align*}
    Var(g_j) &=  Var(K((1-t)^2 - t^2)\mathbf{1}[q = j] ) \\
    &= K(1 - 1/K)((1-t)^2 - t^2)^2 
\end{align*}
For P2 we have:
\begin{align*}
    Var(g_j) &= \mathbb{E}[4K^2 \sum_{i\neq j} \theta_i (1-\theta_i) \mathbf{1}[q = j]] + Var(2*K\sum_{i\neq j} \theta_i \mathbf{1}[q = j] ) \\
    &= 4K\sum_{i\neq j} \theta_i (1-\theta_i) + 4K(1 - 1/K) (\sum_{i\neq j} \theta_i)^2
\end{align*}
after some cancellation. We can likewise compute exact gradients for the DisARM estimator. For P1 we have:
\begin{align*}
    Var(g_j^{\text{DisARM}}) &= E[Var(g_j^{\text{DisARM}})| 1\{z_j \neq \tilde{z}_j\}] + Var[E(g_j^{\text{DisARM}})| 1\{z_j \neq \tilde{z}_j\}] \\
    &= \frac{1 - 2\min(\theta_j, 1-\theta_j)}{2\min(\theta_j,1-\theta_j)} ((1-t)^2 - t^2)^2 + \frac{\sum_{i\neq j} \min(\theta_i, 1-\theta_i)}{\min(\theta_j, 1-\theta_j)} ((1-t)^2 - t^2)^2
\end{align*}
The presence of the second term comes from the increased expected function differences in DisARM due to antithetic sampling. From the observed expressions for P1, it is readily apparent that bitflip variances are lower whenever $K <\frac{1}{2\min(\theta,1-\theta)}$ based on the first term alone. On the other hand when $\theta_i$ becomes large, though the first term is small the second term (representing differences in the function evaluations) becomes much larger. Consider the case $\theta_i = 0.5$ for all $i$. The first term in the variance expression becomes $0$, but the second term in the variance expression becomes larger, in fact $(K-1)((1-t)^2 - t^2)^2$ the exact variance of bitflip-1. In fact, considering $\theta_j \in [0,0.5]$ we see that the derivative of the variance with respect to $\theta_j$ is negative, so for each $\theta_j$ the optimal variance is at $\theta_j = 0.5$. DisARM can thus have lower variance than bitfip-1 when  $\theta_j$ is near $0.5$ but other values $\theta_i$ for $i\neq j$ are near the boundary. We consolidate this into a proposition

\begin{proposition} For P1:
\begin{align*}
    \min_{\theta_1,\dots,\theta_K} \max_{j=1,\dots,K} Var(g_j^{\text{DisARM}}) \geq (K-1) ((1-t)^2 - t^2)^2
\end{align*}
\end{proposition}

\textit{Proof.} Without loss of generality we can replace $1-\theta_j$ with $\theta_j$ whenever $\theta_j > 0.5$. As discussed above, if we have $\theta_1 = \dots = \theta_K$, then we have $\frac{\sum_{i\neq j} \min(\theta_i, 1-\theta_i)}{\min(\theta_j, 1-\theta_j)} ((1-t)^2 - t^2)^2 = (K-1)((1-t)^2 - t^2)^2$ and so the result holds for all $\theta_1 = \dots = \theta_K$. Otherwise choose the largest $\theta_i$ and smallest $\theta_j$ so that $\theta_i > \theta_j$. Then $\sum_{k\neq j} \frac{\theta_k}{\theta_j} > K-1$ since $\frac{\theta_k}{\theta_j} \geq 1$ with strict inequality holding for $i$, which shows the result. This is a special case of Theorem 1. 
\subsection*{Choice of $\tau$ for UGC} We recommend choosing $\tau \geq \frac{1}{2K}$ in all settings, with $\frac{1}{2K}$ guaranteeing lower variance than the family of estimators (Eq. 3-4) containing DisARM and Reinforce-loo as extremes (contingent on assumption 1). In cases where Assumption 1 holds weakly or does not hold (in the sense that $\mathbf{z}$ close to $\tilde{\mathbf{z}}$ being close does not guarantee closeness of  $f(\mathbf{z})$ close to $f(\tilde{\mathbf{z}})$) we recommend using $\frac{1}{2K}$. Such cases include VAEs where $f$ may involve a complex encoder function with no such continuity guarantees. We observe empirically that for VAEs, bitflip-1 gradients are quite high variance when parameter values are far from the boundary (Figure 8-9). On the other hand for functions that (loosely speaking) have such a continuity property such as that of the best subset regression problem, we expect bitflip-1 gradients to have low variance and suggest setting $\tau$ in $(0.1,0.33)$. For this problem we observe robustness to the choice of $\tau$. Future work may formally define classes of functions with varying degress of continuity and analyze optimal estimators for each case.    
\subsection*{P1 Experimental Details} For both P1 and P2, we explore multiple values of $t$ and $K$. For P1 the learning rate is set to $0.8$ and we optimize via projected gradient descent, following standard gradient updates and then clipping the values of the parameters to the range $[0,1]$. We run 1000 iterations of gradient descent, at each iteration computing a gradient variance estimate with $100$ Monte Carlo simulations for each estimator.  Initialization is via standard logistic normal distribution. For UGC, $\tau$ is set to $\frac{1}{2K}$. We report additional results for varying values of $t$ (Figure 10) and $K$ (Figure 11). Variances are clipped to $10,000$ when greater than $10,000$ (only applied to DisARM and Reinforce-loo variance) and smoothed with a moving average of window size $20$. Either log variance or variance is reported and indicated on y axes, depending on which scale gives higher interpretability.  
\subsection*{P2 Experimental Details} To increase variety of settings tested, for P2 we use parameterization by logits $\theta = \frac{e^{\phi}}{ 1+ e^{\phi}}$ and hence no longer have a projection step in gradient descent. We set the learning rate to $2.0$ and initialize each $\theta_j$ to $0.2$ deterministically. For UGC, $\tau$ is set to $\frac{1}{5}$ for all settings tested. Gradient variances are estimated with a $1000$ sample Monte Carlo estimate (and smoothed in the exact same way as P1). We train for $1000$ gradient steps. We report additional results for varying t (Figure 12) and varying K (Figure 13). 
\subsection*{Subset Selection Experimental Details} The number of features is fixed to $200$ , the number of observations is set to $60$ and the number of active features (non-zero coefficients) is set to $3$ (exactly as in \cite{yin2020probabilistic}). The design matrix is sampled $N(0,I)$ for each row and the non-zero $\beta$ are set to $[3,2,1.5]$ (as in \cite{yin2020probabilistic}). $y$ is sampled $N(x^\top \beta, \sigma^2)$, while $\lambda$ and $\sigma^2$ vary throughout the experiments. We train with projected gradient descent for 2000 epochs with learning rate $0.01$ in all experiments. We initialize each coefficient $\hat{\beta}$ at $0.1$ for all experiments. Due to long training times we estimate the gradient variances with a smaller $5$ sample Monte Carlo estimate and apply moving average smoothing, as before. The complete set of results for varying $\lambda$ and $\sigma^2$ are in Figures 14 and 15. UGC is applied with $\tau$ set to $0.33$. The TPR and FPR on held out data are given in Tables 3 and 4. 
\subsection*{Gaussian Mixture Model Experimental Details}

We generate samples from a $20-$dimensional Gaussian mixture model distribution with $6$ components by first sampling component means from a $N(0,8^2)$ distribution, then sampling data conditional on component means from a Normal distribution with variance $\sigma^2$, with $\sigma^2$ being the parameter controlling the signal to noise ratio.  For each of the cluster, $100$ datapoints are sampled. The encoder and decoder architectures are 2-layer neural networks with a $25$-dimensional intermediate layer, with Relu nonlinearities and $0.1$ dropout. Optimization is via Adam optimizer with learning rate $0.01$, trained for $1000$ epochs. Gradient variances are estimated with a $100-$sample Monte Carlo estimate at the end of every epoch. Variances are clipped at $10e7$ (only for DisARM and Reinforce-loo) before a moving average is applied with window size $30$. 

\subsection*{Results on variational autoencoders: FashionMNIST,DynamicMNIST and Omniglot}
We repeat the experiment on DynamicMNIST, FashionMNIST and Omniglot of \cite{dong2020disarm} with latent dimension set to $30$ and random normal initialization of all parameters $N(0,0.3^2)$. We train with learning rates of $1e-3$ for the encoder and decoder and $1e-2$ for the prior variables. The detailed description of model and experiment can be found in \cite{dong2020disarm}. The results across all settings tested appear in Figure 8 and Figure 9. 

\begin{figure*}[!ht]
         \centering
         \includegraphics[width=\textwidth]{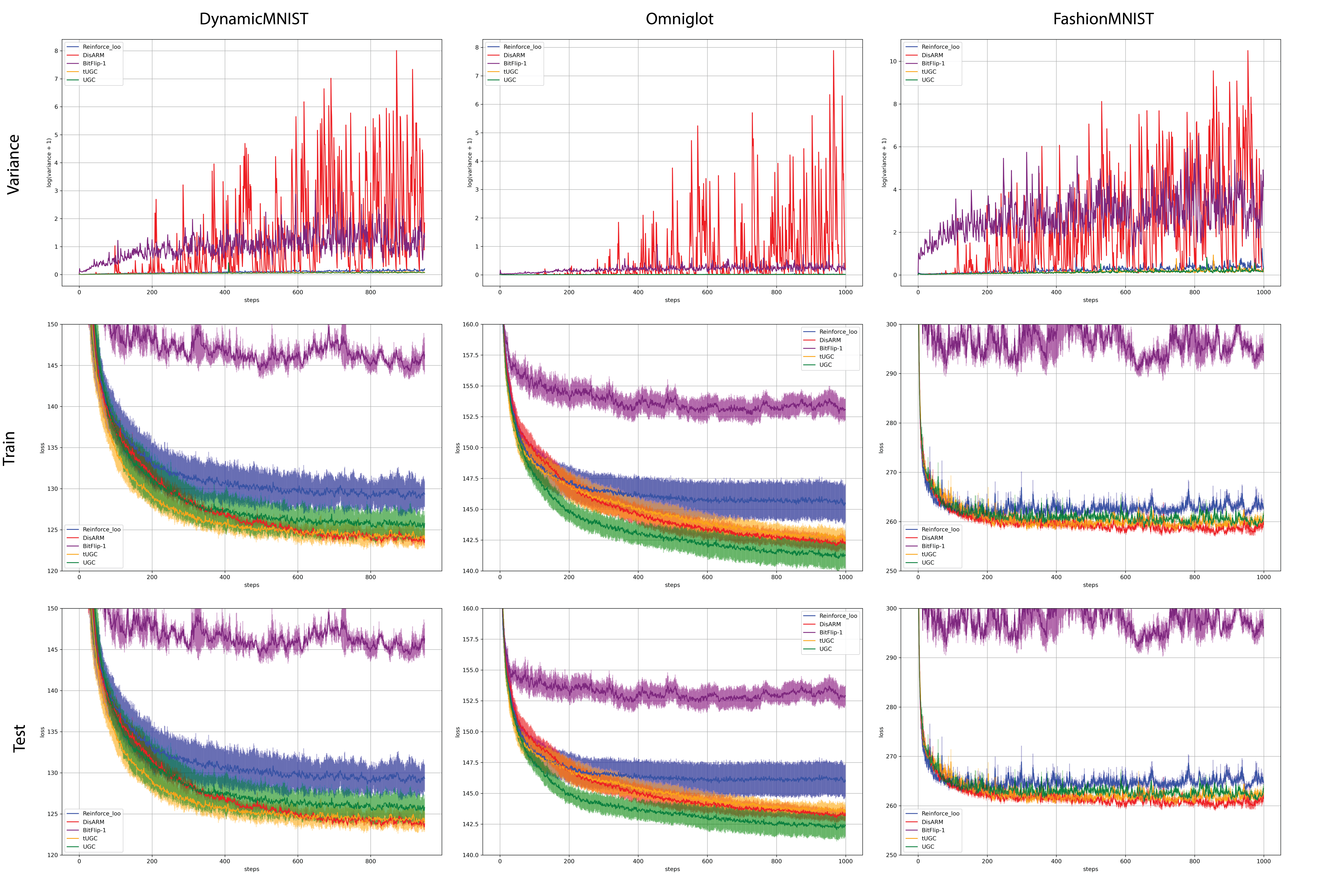}
     \caption{Experiment on VAE with  nonlinear encoder and decoder  on Dynamic MNIST, FashionMNIST, and Omniglot datasets}
\end{figure*}

\begin{figure*}[!ht]
         \centering
         \includegraphics[width=\textwidth]{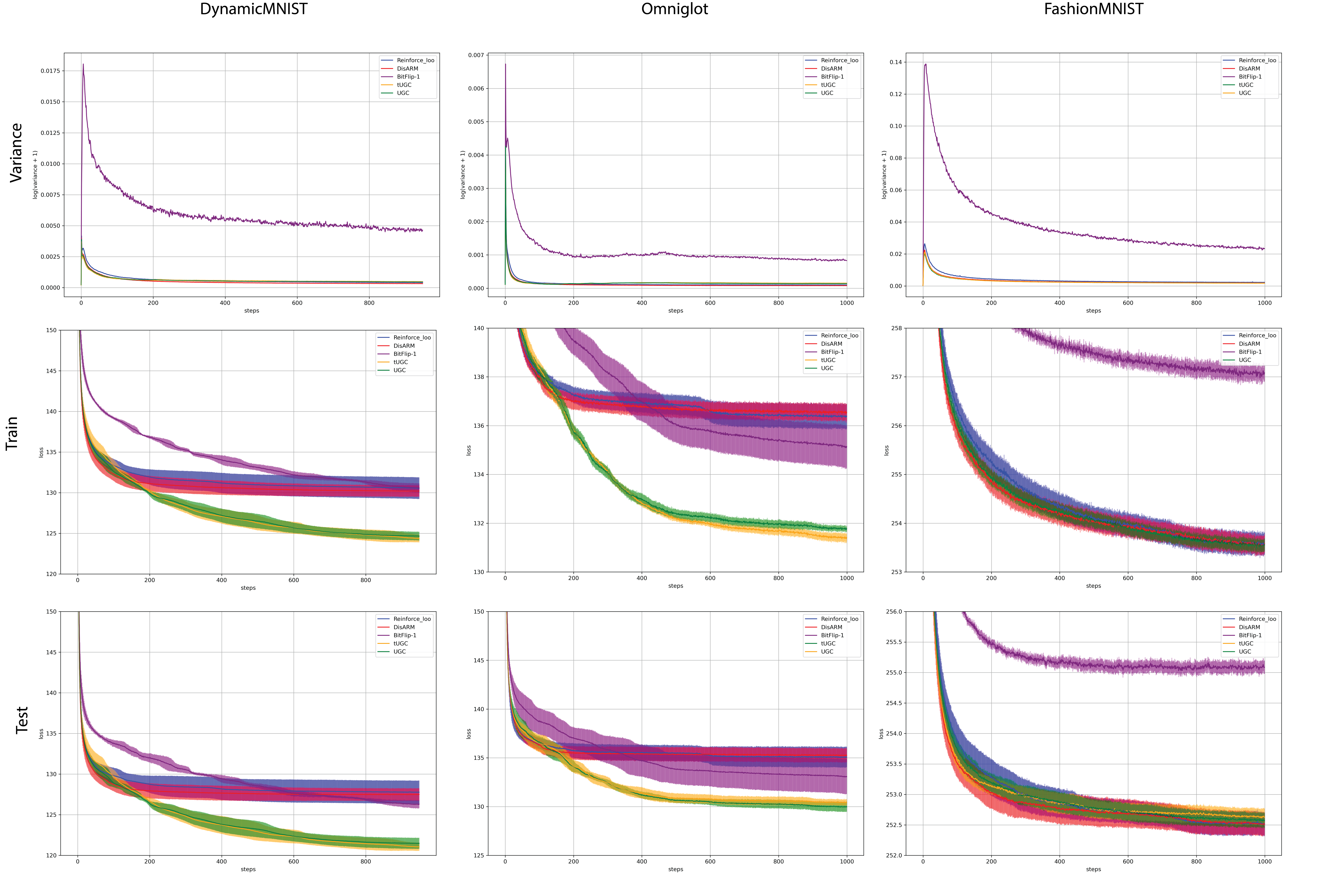}
     \caption{Experiment on VAE with  linear encoder and decoder  on Dynamic MNIST, FashionMNIST, and Omniglot datasets}
\end{figure*}


\begin{figure*}[!ht]
         \centering
         \includegraphics[width=\textwidth]{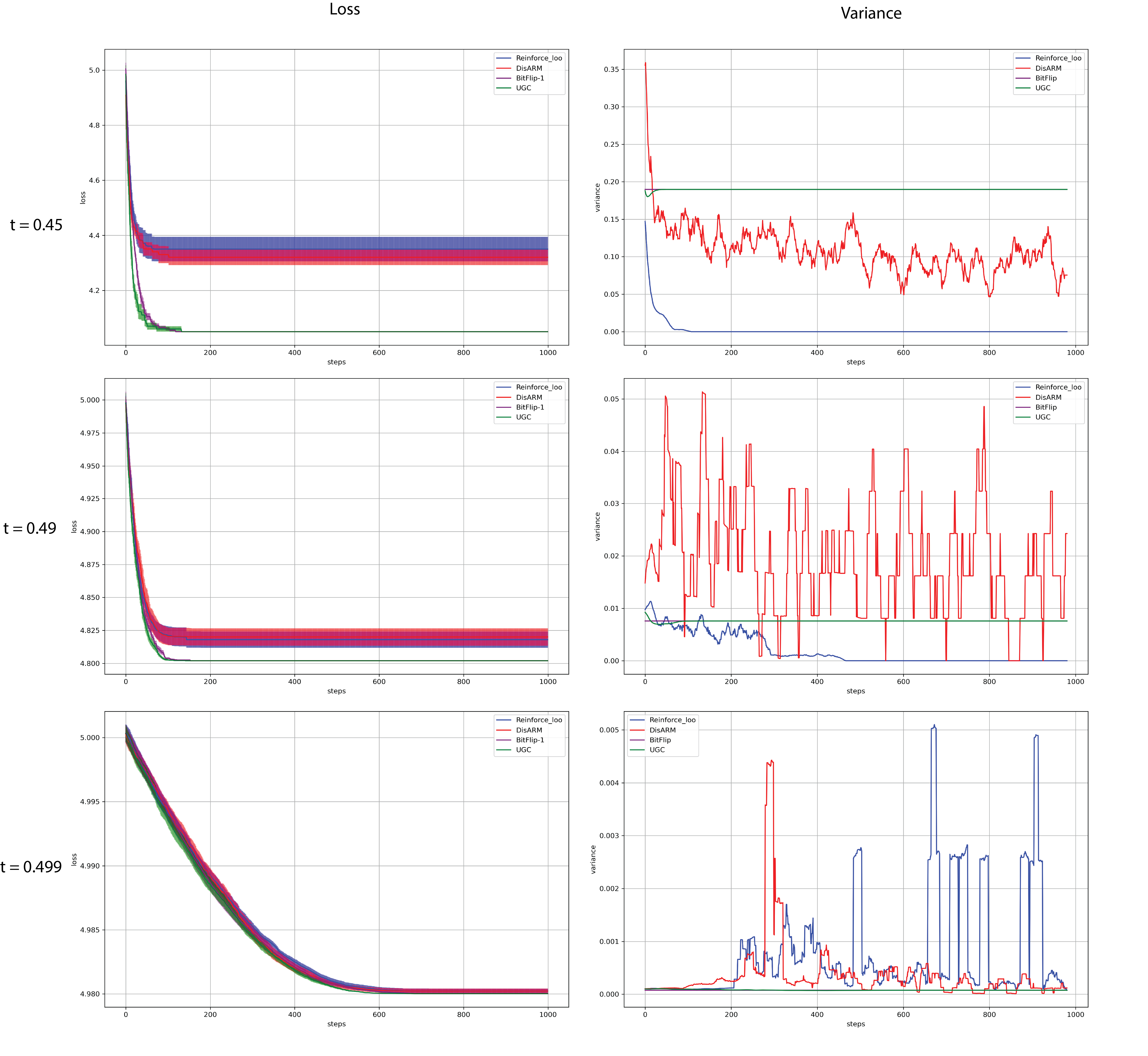}
     \caption{Additional experiments for P1 with varying values of $t$}
\end{figure*}

\begin{figure*}[!ht]
         \centering
         \includegraphics[width=\textwidth]{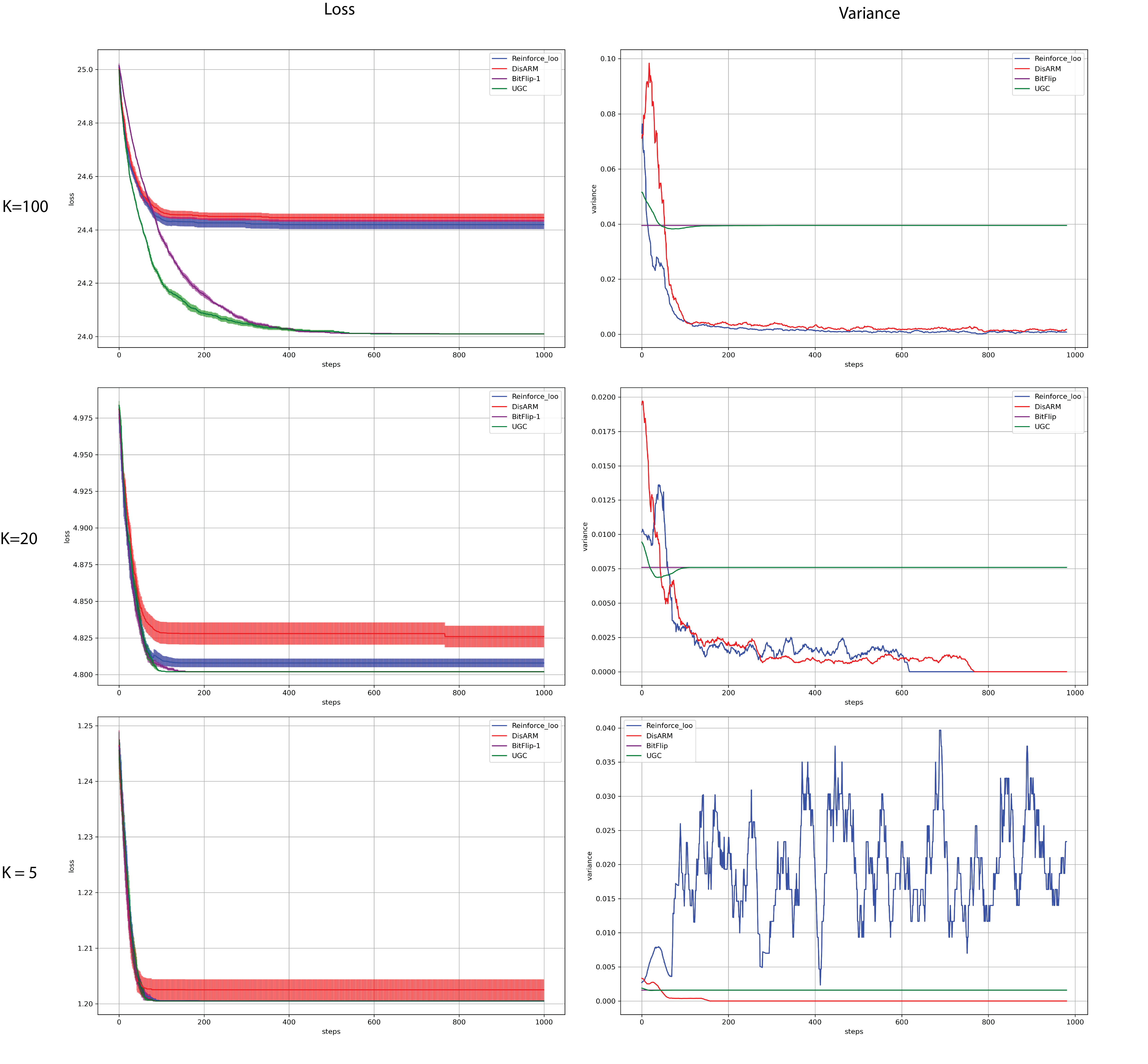}
     \caption{Additional experiments for P1 with varying values of $K$}
\end{figure*}

\begin{figure*}[!ht]
         \centering
         \includegraphics[width=\textwidth]{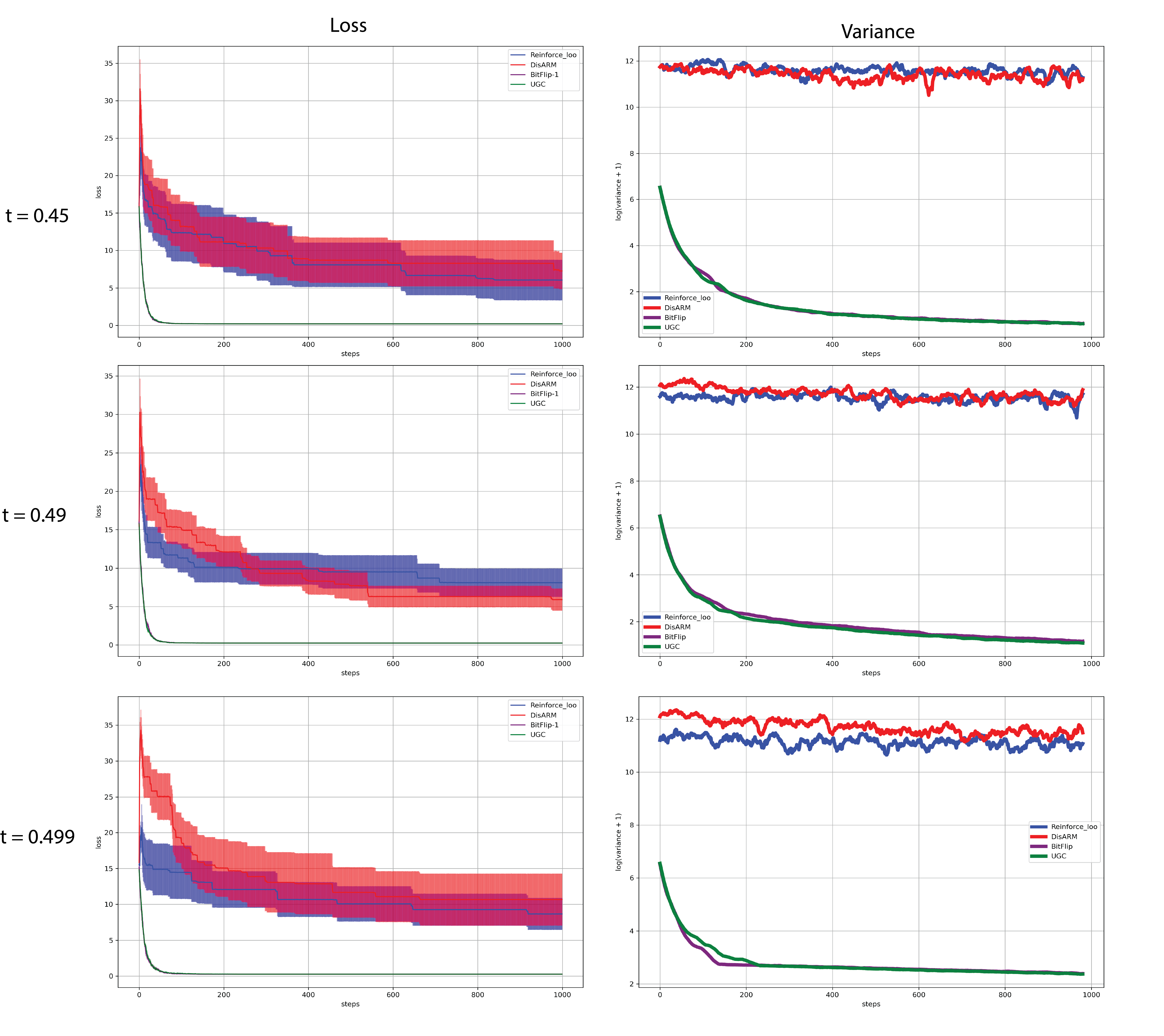}
     \caption{Additional experiments for P2 with varying values of $t$}
\end{figure*}

\begin{figure*}[!ht]
         \centering
         \includegraphics[width=\textwidth]{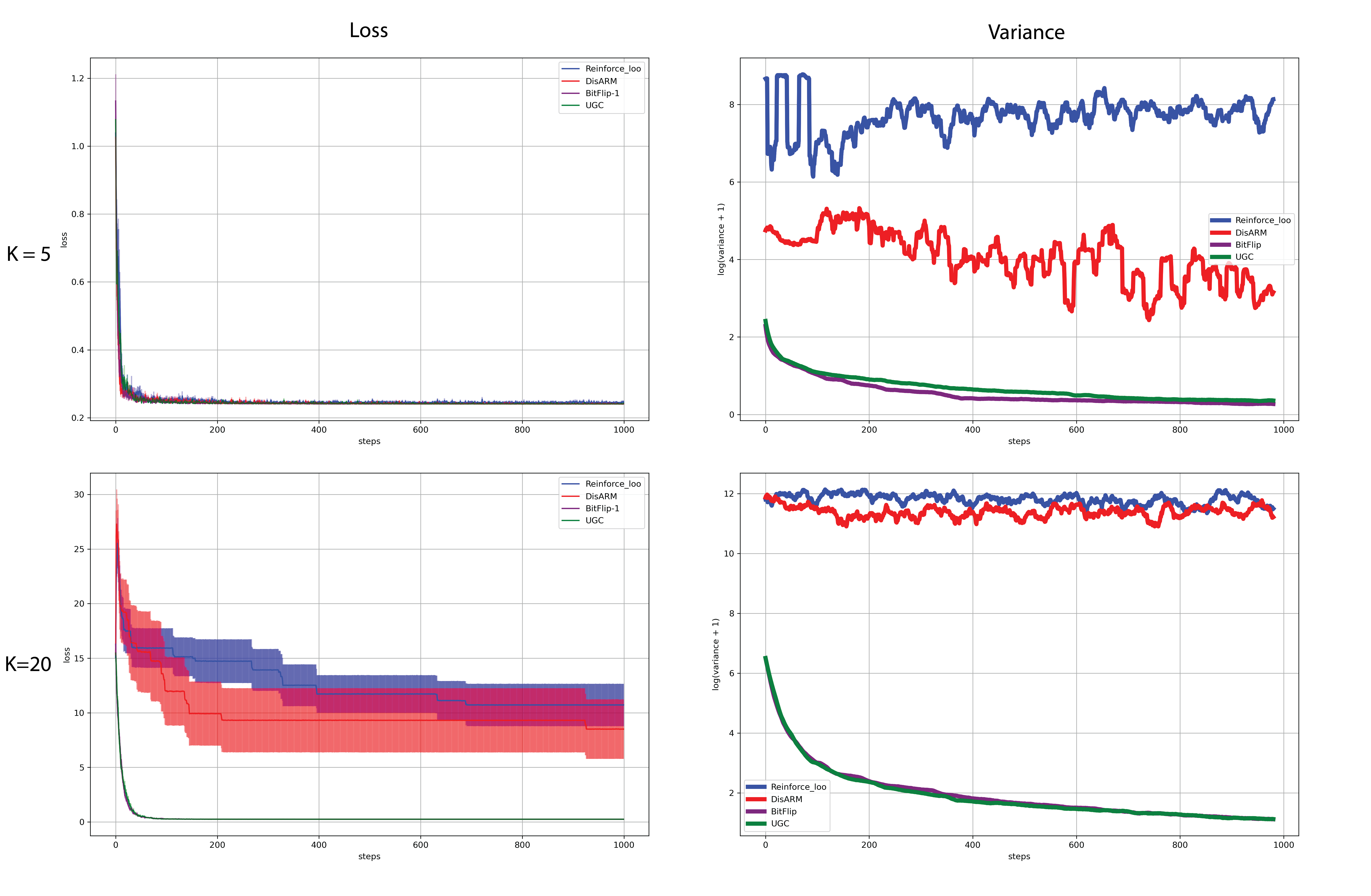}
     \caption{Additional experiments for P2 with varying values of $K$}
\end{figure*}

\newpage 
\cheading{Subset selection accuracy for threshold $\theta > 0.5$ }{$\lambda = 1$}

\begin{longtable}{@{}l rr rr rr rr rr rr rr rr }

\toprule%
 \centering%
 
 & \multicolumn{4}{c}{{{\bfseries bitflip-1}}}
 & \multicolumn{4}{c}{{{\bfseries UGC}}} 
 & \multicolumn{4}{c}{{{\bfseries DisARM}}} 
 & \multicolumn{4}{c}{{{\bfseries Reinforce-loo}}} \\

 \centering%
 \bfseries SNR
 & \multicolumn{2}{c}{{{\bfseries TPR}}}
 & \multicolumn{2}{c}{{{\bfseries FPR}}}
 & \multicolumn{2}{c}{{{\bfseries TPR}}}
 & \multicolumn{2}{c}{{{\bfseries FPR}}}
 & \multicolumn{2}{c}{{{\bfseries TPR}}}
 & \multicolumn{2}{c}{{{\bfseries FPR}}}
 & \multicolumn{2}{c}{{{\bfseries TPR}}}
 & \multicolumn{2}{c}{{{\bfseries FPR}}}\\

\cmidrule[0.4pt](r{0.125em}){1-1}%
\cmidrule[0.4pt](lr{0.125em}){2-3}%
\cmidrule[0.4pt](lr{0.125em}){4-5}%
\cmidrule[0.4pt](lr{0.125em}){6-7}%
\cmidrule[0.4pt](lr{0.125em}){8-9}%
\cmidrule[0.4pt](lr{0.125em}){10-11}%
\cmidrule[0.4pt](lr{0.125em}){12-13}%
\cmidrule[0.4pt](lr{0.125em}){14-15}%
\cmidrule[0.4pt](lr{0.125em}){16-17}%
\endhead

$15.25$ & \highest{0.96} & \highest{(0.1)} & \highest{0.0} &
\highest{(0.0)} & \highest{0.96} & \highest{(0.1)} & \highest{0.0} &
\highest{(0.0)} & 0.56 & (0.26) & 0.05 & (0.02) & 0.6 & (0.36) & 0.04 & (0.01)\\

\myrowcolour%
$3.81$  & \highest{1.0} & \highest{(0.0)} & \highest{0.0} &
\highest{(0.0)} & \highest{1.0} & \highest{(0.0)} & \highest{0.0} &
\highest{(0.0)} & 0.66 & (0.26) & 0.06 & (0.03) & 0.53 & (0.16) & 0.04 & (0.01)\\

$1.69$ & 0.83 & (0.27) & \highest{0.01} &
\highest{(0.01)} & \highest{0.87} & \highest{(0.22)} & \highest{0.01} &
\highest{(0.01)} & 0.43 & (0.26) & 0.06 & (0.03) & 0.50 & (0.31) & 0.05 & (0.02)\\

\myrowcolour%
$0.95$ & 0.43 & (0.30) & 0.04 &
(0.01) & \highest{0.67} & \highest{(0.21)} & \highest{0.03} &
\highest{(0.01)} & 0.40 & (0.29) & 0.06 & (0.02) & 0.43 & (0.21) & 0.05 & (0.01)\\

\bottomrule

\caption{Results on projected gradient descent best subset regression problem for varying signal to noise ratios (SNR), defined as $\beta^\top \beta / \sigma^2$. Rounded to 2 decimal places. The number of features $p=200$, while the active set of features is size $3$. The number of observations $n =60$. UGC achieves the highest performance across settings}
\end{longtable}

\begin{figure*}[!ht]
         \centering
         \includegraphics[width=\textwidth]{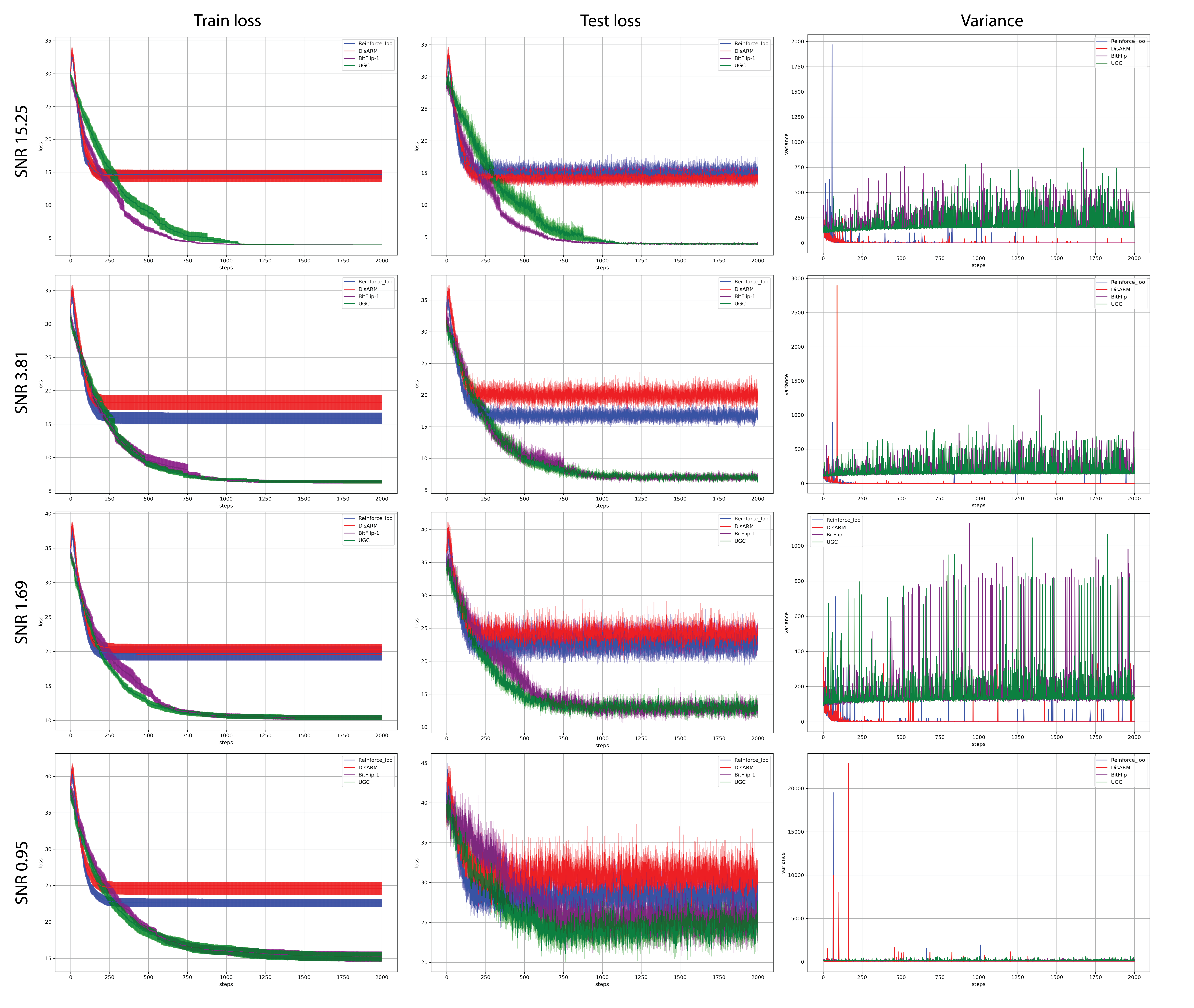}
     \caption{Additional experiments for the best subset selection linear regression problem for varying levels of signal to noise ratio; using projected gradient descent. Low signal to noise ratio generally does not affect convergence (left plots) but affects solution quality as measured by the loss on newly simulated data (middle). On the right, we see that DisARM and Reinforce-loo experience high gradients early in training; however, quickly converge to the wrong solution, at which point they become stuck and thereafter have low variance gradients}
\end{figure*}
\newpage
\cheading{Subset selection accuracy for threshold $\theta > 0.5$ }{SNR = $3.81$}

\begin{longtable}{@{}l rr rr rr rr rr rr rr rr }

\toprule%
 \centering%
 
 & \multicolumn{4}{c}{{{\bfseries bitflip-1}}}
 & \multicolumn{4}{c}{{{\bfseries UGC}}} 
 & \multicolumn{4}{c}{{{\bfseries DisARM}}} 
 & \multicolumn{4}{c}{{{\bfseries Reinforce-loo}}} \\

 \centering%
 \bfseries $\lambda$
 & \multicolumn{2}{c}{{{\bfseries TPR}}}
 & \multicolumn{2}{c}{{{\bfseries FPR}}}
 & \multicolumn{2}{c}{{{\bfseries TPR}}}
 & \multicolumn{2}{c}{{{\bfseries FPR}}}
 & \multicolumn{2}{c}{{{\bfseries TPR}}}
 & \multicolumn{2}{c}{{{\bfseries FPR}}}
 & \multicolumn{2}{c}{{{\bfseries TPR}}}
 & \multicolumn{2}{c}{{{\bfseries FPR}}}\\

\cmidrule[0.4pt](r{0.125em}){1-1}%
\cmidrule[0.4pt](lr{0.125em}){2-3}%
\cmidrule[0.4pt](lr{0.125em}){4-5}%
\cmidrule[0.4pt](lr{0.125em}){6-7}%
\cmidrule[0.4pt](lr{0.125em}){8-9}%
\cmidrule[0.4pt](lr{0.125em}){10-11}%
\cmidrule[0.4pt](lr{0.125em}){12-13}%
\cmidrule[0.4pt](lr{0.125em}){14-15}%
\cmidrule[0.4pt](lr{0.125em}){16-17}%
\endhead

$0.01$ & 0.30 & (0.31) & \highest{0.31} &
\highest{(0.01)} &  0.53 & (0.27) & 0.34 &
(0.03) & \highest{0.83} & \highest{(0.22)} & 0.34 & (0.02) & \highest{0.83} & \highest{(0.17)} & \highest{0.31} & \highest{(0.03)}\\

\myrowcolour%
$0.1$  & 0.73 & (0.29) & 0.11 &
(0.07) & \highest{0.97} & \highest{(0.10)} & 0.10 &
(0.06) & 0.83 & (0.17) & \highest{0.08} & \highest{(0.03)} & 0.87 & (0.16) & \highest{0.08} & \highest{(0.03)}\\

$1.0$ & \highest{0.93} & \highest{(0.13)} & \highest{0.0} &
\highest{(0.0)} & \highest{0.93} & \highest{(0.13)} & \highest{0.0} &
\highest{(0.0)} & 0.47 & (0.27) & 0.06 & (0.02) & 0.60 & (0.33) & 0.05 & (0.02)\\

\myrowcolour%
$10.0$ & 0.10 & (0.15) & \highest{0.0} &
\highest{(0.0)} & 0.10 & (0.15) & \highest{0.0} &
\highest{(0.0)} & \highest{0.50} & \highest{(0.31)} & 0.30 & (0.03) & 0.23 & (0.26) & 0.33 & (0.01)\\

\bottomrule

\caption{Results on projected gradient descent best subset regression problem for varying $\lambda$. Rounded to 2 decimal places. The number of features $p=200$, while the active set of features is size $3$. The number of observations $n =60$. UGC achieves the highest performance across settings}
\end{longtable}

\begin{figure*}[!ht]
         \centering
         \includegraphics[width=\textwidth]{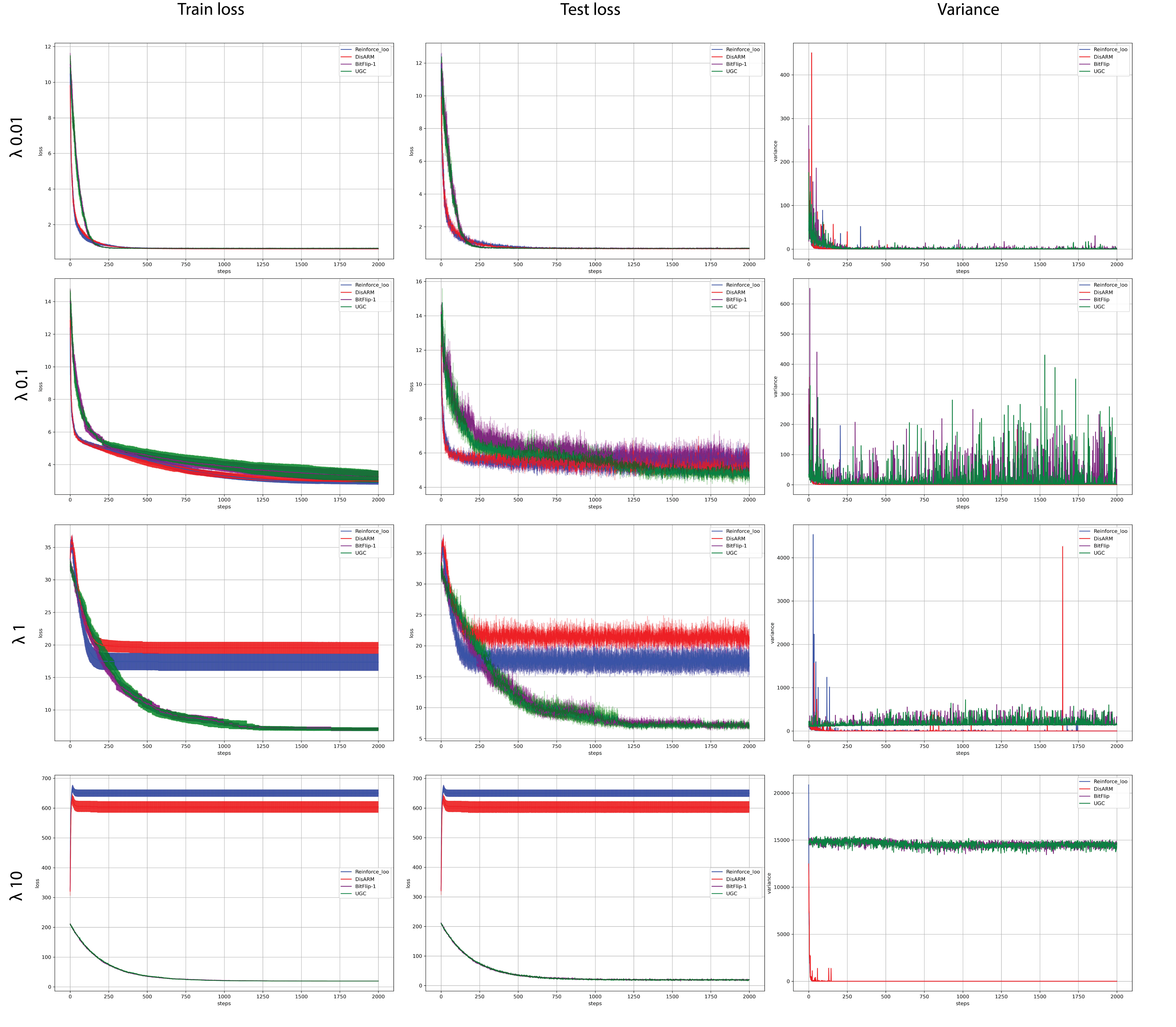}
     \caption{Additional experiments for the best subset selection linear regression problem for varying levels of tuning parameter $\lambda$; using projected gradient descent. At low values of $\lambda$ all methods converge, with DisARM and Reinforce-loo performing favorably. However at high values of $\lambda$ DisARM and Reinforce-loo converge to an incorrect solution (bottom rows). This reinforces the idea that the estimators have different behaviors for different functions}
\end{figure*}

\end{document}